\documentclass[lettersize,journal,twoside]{IEEEtran}
\usepackage{amsmath,amsfonts,amssymb}
\usepackage{algorithmic}
\usepackage{algorithm}
\usepackage{array}
\usepackage[caption=false,font=normalsize,labelfont=sf,textfont=sf]{subfig}
\usepackage{textcomp}
\usepackage{stfloats}
\usepackage{url}
\usepackage{verbatim}
\usepackage{graphicx}
\usepackage{cite}

\newcommand{\etal}{\textit{et al}.}
\newcommand{\ie}{\textit{i}.\textit{e}.}
\newcommand{\eg}{\textit{e}.\textit{g}.}
\newcommand{\etc}{\textit{etc}}

\usepackage{enumerate}
\usepackage{hyperref}
\usepackage{tabularx}
\usepackage{multirow,multicol}
\usepackage[table]{xcolor}
\usepackage{url}
\usepackage{color}
\usepackage{soul}
\usepackage{makecell}

\begin{document}

\title{Understanding and Mitigating Overfitting in \\Prompt Tuning for Vision-Language Models}

\author{Chengcheng Ma$^\ast$, Yang Liu, Jiankang Deng,~\IEEEmembership{Student Member,~IEEE}, Lingxi Xie, \\Weiming Dong$^\dag$,~\IEEEmembership{Member,~IEEE}, Changsheng Xu,~\IEEEmembership{Fellow,~IEEE}
\thanks{$^\ast$This work was done during an internship in Huawei Inc. }
\thanks{Chengcheng Ma, Weiming Dong, and Changsheng Xu are with National Lab of Pattern Recognition (NLPR), Institute of Automation, Chinese Academy of Sciences (CASIA), Beijing, 100190, China.}
\thanks{Chengcheng Ma is also with School of Artificial Intelligence, University of Chinese Academy of Sciences (UCAS), Beijing, 100049, China.}
\thanks{Yang Liu is with Alibaba DAMO Academy, Hangzhou, 310024, China.}
\thanks{JianKang Deng and Lingxi Xie are with Huawei Inc., Shenzhen
518129, China.}
\thanks{$^\dag$Corresponding author (weiming.dong@ia.ac.cn).}
}



\maketitle

\begin{abstract}
Pretrained vision-language models (VLMs) such as CLIP have shown impressive generalization capability 
in downstream vision tasks with appropriate text prompts.
Instead of designing prompts manually, Context Optimization (CoOp) has been recently proposed to learn continuous prompts using task-specific training data.
Despite the performance improvements on downstream tasks, 
several studies have reported that CoOp suffers from the overfitting issue in two aspects: 
(i) the test accuracy on base classes first improves and then worsens during training;
(ii) the test accuracy on novel classes keeps decreasing. 
However, none of the existing studies can 
understand and mitigate such overfitting problems. 
%
%
In this study, we first explore the cause of overfitting by analyzing the gradient flow.
Comparative experiments reveal that CoOp favors generalizable and spurious features in the early and later training stages, respectively, leading to the non-overfitting and overfitting phenomena.
Given those observations, we propose Subspace Prompt Tuning (\textit{Sub}PT) to project the gradients in back-propagation onto the low-rank subspace spanned by the early-stage gradient flow eigenvectors during the entire training process and successfully eliminate the overfitting problem. 
In addition, we equip CoOp with a Novel Feature Learner (NFL) to enhance the generalization ability of the learned prompts onto novel categories beyond the training set, needless of image training data.  
Extensive experiments on 11 classification datasets demonstrate that \textit{Sub}PT+NFL consistently boost the performance of CoOp and outperform the state-of-the-art CoCoOp approach.
Experiments on more challenging vision downstream tasks, including open-vocabulary object detection and zero-shot semantic segmentation, also verify the effectiveness of the proposed method.
Codes can be found at \url{https://tinyurl.com/mpe64f89}.
\end{abstract}




\section{Introduction}
\label{sec: intro}
\IEEEPARstart{W}{ith} the recent advances in vision-language pretraining, such as CLIP \cite{CLIP} and ALIGN \cite{ALIGN}, there has been a growing interest in developing multimodal foundation models for downstream tasks \cite{FoundationModels,mei2022guest,zhang2020language}.
These vision-language models (VLMs) are pretrained on millions of image-text data pairs to align the vision and language modalities in the same feature space, and the resulting models can obtain impressive performances on downstream tasks in a zero-shot manner with a properly designed text prompt only, instead of any task-related training data.
Early works such as \cite{CLIP} and \cite{ALIGN} adopted handcrafted prompt templates like ``a photo of a [CLASS]”, which usually lack task-specific heuristics and are non-robust across different target domains. 
\begin{figure}[t]
    \centering
    \scalebox{0.55}{
    \includegraphics{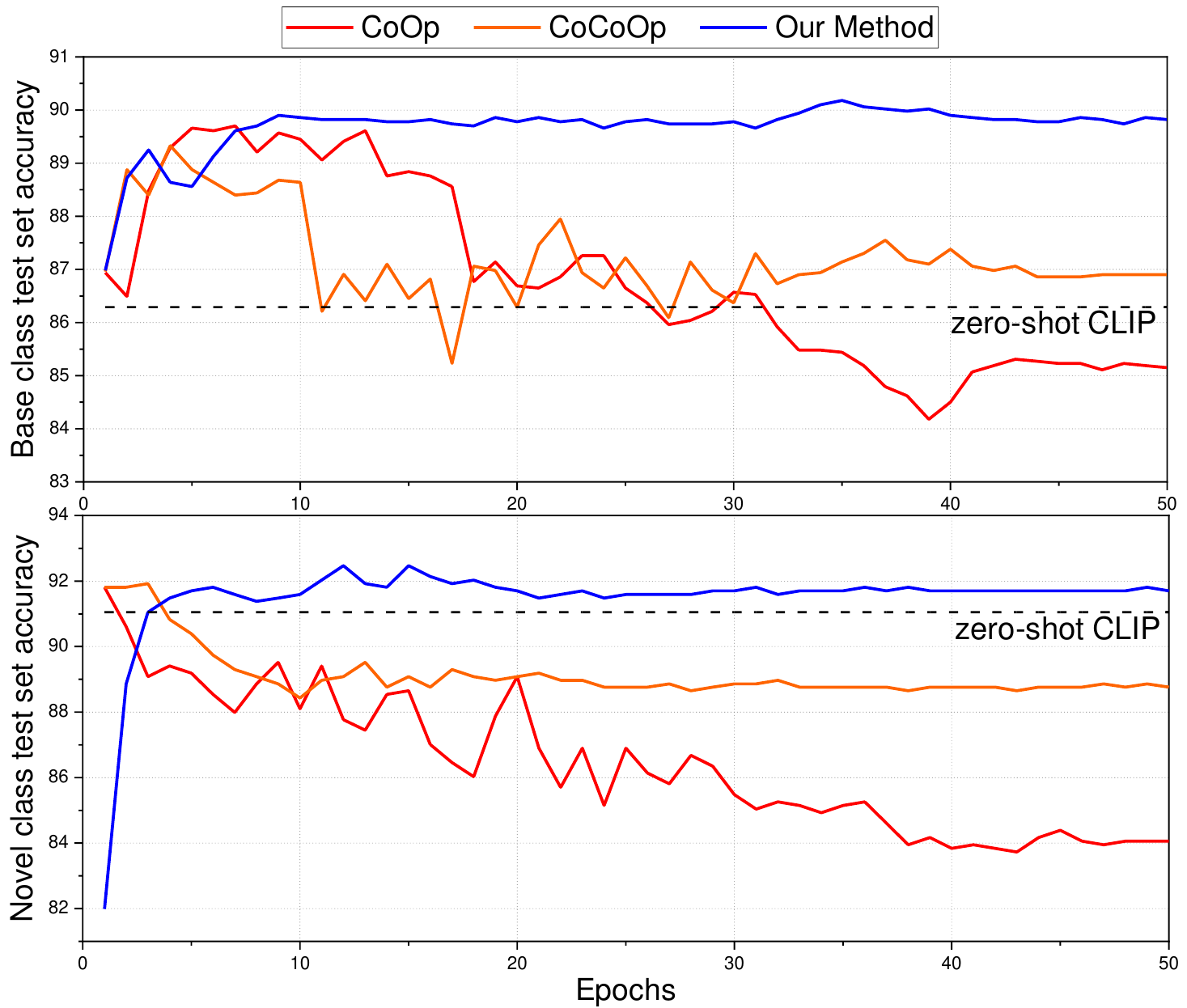}
    }
    \caption{\textbf{Illustrations of two aspects of the overfitting issue in CoOp and CoCoOp}. The dataset is Caltech101. 
    (a) \textbf{Top}: The test accuracy on base classes first improves and then worsens during the training processes of CoOp and CoCoOp. 
    (b) \textbf{Bottom}: The test accuracy on novel classes keeps decreasing and become far below that of the zero-shot CLIP. 
    Our proposed \textit{Sub}PT and NFL can successfully mitigate the overfitting problem.
    }
    \label{fig: base and novel}
\end{figure}
%
Inspired by the recent works in the NLP community \cite{lester2021power,li2021prefix}, Zhou \etal\cite{CoOp} proposed a prompt tuning method named Context Optimization (CoOp) to address this problem, in which the prompt can be optimized in a continuous space. 
CoOp can learn machine-favorable prompts and significantly improve CLIP's performance on the classification task compared with handcrafted prompts. 

Although effective, 
some recent studies such as \cite{CoCoOp,zhu2022prompt,derakhshani2022variational} found that CoOp suffers from the overfitting issue in two aspects. 
First, Zhu \etal\cite{zhu2022prompt} found that the test accuracy on base classes first improves and then worsens, and heavily drops by at most 4\% when training ends. We also observe such common failure across many datasets in our experiments (see Fig. \ref{fig: base and novel}(a) for an example). 
Second, Zhou \etal\cite{CoCoOp} found that the learned prompt is not applicable to novel classes beyond the training set, indicating that CoOp hurts the generalization capability of CLIP. 
As displayed in Fig. \ref{fig: base and novel}(b), the test accuracy on novel classes keeps decreasing during training, far below that of the zero-shot CLIP baseline with the handcrafted prompt.

Unfortunately, none of the existing works can effectively mitigate the overfitting problem in CoOp. 
For instance, Zhou \etal\cite{CoCoOp} introduced Conditional CoOp (CoCoOp) and designed image-specific prompts by adding prompt embedding 
with MLP-transformed image features. 
However, as shown in Figs.~\ref{fig: base and novel}(a) and (b), CoCoOp \cite{CoCoOp} still fails to prevent base class test set accuracy from decreasing in the later training stage, and the novel class test accuracy is still below that of the zero-shot CLIP baseline.
Furthermore, we notice that the conventional anti-overfitting strategies, \eg, early stopping and data augmentation, do not work in CoOp. 
Concretely, CoOp is based on a few-shot setting, so collecting a validation set for applying early stopping is unrealistic. Besides, data augmentation is not always robust across various downstream datasets and tasks.
Additionally, overfitting cannot be detected during the entire training process, because the accuracy/loss on training set always keeps increasing/decreasing. 
This situation suggests the need to first understand the cause of overfitting and then address the problem accordingly.

In this study, we first manage to understand the cause of overfitting in CoOp. 
According to the no free lunch theorem \cite{NoFreeLunch,GradientStarvation}, an intuitive understanding is that CoOp favors the generalizable/spurious feature in the early/later training stage, leading to an increase/decrease in the test set accuracy.
To investigate the learned features, we measure the gradient flows in early and later training stages, and utilize the dominant eigenvectors of the corresponding checkpoint trajectories to approximate the gradient flow for computational convenience.
We observe 
the same phenomenon across several datasets simultaneously 
in that the two eigenvectors representing early-stage and later-stage gradient flows are almost orthogonal, revealing that CoOp does favor different feature directions in different training stages. 
To further identify the effects of these two feature directions, 
we conduct comparative experiments to rerun CoOp twice from the same initial point, and project the gradients in back-propagation onto the subspace spanned by the early-stage and later-stage eigenvectors during the whole training process, respectively.
The large performance gap demonstrates that CoOp 
indeed favors the generalizable/spurious feature in the early/later training stage. 

On the basis of the aforementioned observations, we propose Subspace Prompt Tuning (\textit{Sub}PT) to mitigate the overfitting problem in prompt tuning.
In particular, we first conduct CoOp and collect the saved checkpoints within the early training stage to compute the dominant eigenvectors.
Then we rerun CoOp from the same initial point and project the gradients onto the subspace spanned by the precomputed early-stage eigenvectors during the entire training process.
The rationale behind \textit{Sub}PT lies in that the spurious components in gradients can be eliminated by projection since they are orthogonal to the early-stage gradient flow. 
Moreover, we design a Novel Feature Learner (NFL) to enhance the generalization capability towards novel classes with no need for any image training data.
NFL maximizes the similarity between text features generated from learnable prompt embeddings and an ensemble of handcrafted prompts.
Extensive experiments on three downstream vision tasks, \ie, image classification, open-vocabulary object detection, and zero-shot semantic segmentation, verify the effectiveness of the proposed approach.

The contributions of this work are summarized as follows:
\begin{itemize}
    \item We manage to analyze the cause of the overfitting problem in CoOp by measuring the gradient flow in the early and later training stage.
    \item We propose Subspace Prompt Tuning (\textit{Sub}PT) to eliminate the spurious components in gradients during back-propagation and successfully mitigate overfitting.
    \item We propose Novel Feature Learner (NFL) to enhance the generalization ability of the learned prompts towards novel categories with no need for any image training data.
    \item Extensive experiments on three downstream vision tasks, including image classification, open-vocabulary object detection, and zero-shot semantic segmentation, verify the effectiveness of the proposed approach.
\end{itemize}

\section{Related Work}
In this section, we briefly revisit the vision-language models and existing prompt tuning methods. Then we review the representative studies on the generalization under data distribution shifts between the training and test set.

\subsection{Vision-Language Models}
During the last decades, vision-language models (VLMs) \cite{yu2019multimodal,zhang2020language,yang2020grounding,mei2022guest} have received great research attention because they show more potential for understanding real-world data and activities than unimodal learning \cite{cheng2022hybrid}.
Limited by costly training data, previous VLMs can usually handle only one task and one dataset, showing less zero-shot capabilities.
Inspiringly, CLIP \cite{CLIP}, pretrained on 400 million image-caption pairs scratched from the web, has been recently proposed to address this problem.
By design, CLIP consists of an image encoder and a text encoder, and can be applied to any image classification dataset by simply filling the names of categories to be recognized into a properly designed prompt template, such as ``a photo of a [CLASS]".
Other VLMs, such as ALIGN \cite{ALIGN} and LiT \cite{zhai2022lit}, 
have been concurrently proposed towards the same goal and can be extended to more challenging visual recognition tasks 
\cite{liang2022visual,tsimpoukelli2021multimodal,wang2022ofa,liang2022local,wang2022counterfactual}.
%
This research is orthogonal to the aforementioned works, aiming to enhance the adaptation of CLIP to downstream tasks.

\subsection{Prompt Tuning for Vision-Language Models}

Many recent works have explored efficient and effective approaches for adapting VLMs to downstream vision tasks.
%
As a pioneer work, Zhou \etal \cite{CoOp} were inspired by a recent advance in the NLP field \cite{lester2021power} and proposed Context Optimization (CoOp) to improve the performance of CLIP towards the downstream image classification task.
Instead of using handcrafted prompt templates, CoOp is introduced to learn the machine-favorable prompt embeddings with only a few training samples from the downstream dataset.
Following CoOp, some other prompt tuning methods \cite{lu2022prompt,xing2022class,yao2022pevl} have also been proposed towards the same goal.
However, some recent studies \cite{CoCoOp,zhu2022prompt,derakhshani2022variational} report that the CoOp training suffers from severe overfitting issue, which hurts the generalization ability on novel categories and out-of-distribution data.
To address this issue, Conditional CoOp (CoCoOp) \cite{CoCoOp} presents image-specific prompt tuning by introducing a Meta-Net. 
ProGrad \cite{zhu2022prompt} follows PCGrad \cite{PCGrad} to utilize the supervision signal from zero-shot CLIP to conduct gradient surgery during training. 
VPT \cite{derakhshani2022variational} upgrades CoCoOp with the variational framework.
However, all of the aforementioned works increase the computational burden by introducing additional network components, and fail to understand the cause of overfitting issue. 
Thus, how to mitigate the overfitting remains an open question.

Besides, we note that some other works aim at studying vision prompts, such as tip-adapter \cite{tip-adapter}, colorful prompt tuning \cite{CPT}, visual prompt tuning \cite{VPT}, and neural prompt search \cite{NPS}.
However, the vision modality is naturally more complex than the language modality due to the diversity of illumination, viewpoint, domain, \etc., suggesting that vision prompts may be less generalizable than text prompts across tasks and datasets.
Thus, we only study prompt tuning on the text branch in this work.

\subsection{Generalization under Data Distribution Shifts}
In the machine learning community, the conventional goal is to minimize the total error on the test set, which shares an identical distribution and is independent of the training set.
However, such an i.i.d assumption is usually unsatisfactory in real-world scenario due to changes in the environment \cite{wilds} and the occurrence of novel categories \cite{openset}, and the data distribution shifts refer to the mismatch of distribution between training and test data.
How to train reliable and robust models under data distribution shifts has been studied differently in various research areas.
For domain generalization \cite{domain-generalization,domain-generalization2,wang2022feature}, researchers focus on learning domain-invariant features across training datasets from multiple source domains.
For meta-learning and few-shot learning \cite{free,jiang2020multi,cheng2021meta}, researchers have access to a large training set to learn a similarity function, and conduct inference on the query set with the help of a small support set.
However, prompt tuning is different from all mentioned works because only a few training samples from a single domain are provided, and only prompt embeddings (classifier) rather than visual and text encoders (feature extractor) are trainable.

\section{Preliminary}
\subsection{Contrastive Language-Image Pre-training (CLIP)}
CLIP \cite{CLIP} is composed of two independent encoders, in which the visual encoder $f(\cdot)$ maps the input image $\mathbf{x}$ into the feature vector $f(\mathbf{x})$, and the text encoder $g(\cdot)$ maps the input sentence $\mathbf{t}$ into $g(\mathbf{t})$. For clarity, we denote the $\ell_2$-normalized feature as $\mathbf{z}=\frac{f(\mathbf{x})}{\left \| f(\mathbf{x}) \right \|}$ and $\mathbf{w}=\frac{g(\mathbf{t})}{\left \| g(\mathbf{t}) \right \|}$ sharing the same dimensionality. 
The inner product $\mathbf{z}^\intercal\mathbf{w}$ stands for the semantic similarity between $\mathbf{x}$ and $\mathbf{t}$.

We revisit the method of utilizing CLIP on zero-shot image classification. 
In the baseline setting, the prompt is manually designed as ``a photo of a [CLASS]". 
Given a $C$-way task, we first prepare $C$ category descriptions by feeding each of the category names (already known) into the prompt, and obtain $C$ text features $\{\mathbf{w}_i\}_{i=1}^C$ using the text encoder $g(\cdot)$. Then, given an input image $\mathbf{x}$, we compute similarity scores between the encoded image feature $\mathbf{z}$ and all text features, and finally obtain the prediction probability on class $y$ via the softmax function as
\begin{equation}
\label{eq: clip}
p(y|\mathbf{x})=\frac{\exp(\mathbf{z}^\intercal\mathbf{w}_y/\tau)}{\sum_{i=1}^{C}\exp(\mathbf{z}^\intercal\mathbf{w}_i/\tau)}.
\end{equation}
We omit the temperature $\tau$ (default as 1.0) in the following sections for simplicity.
As CLIP is pretrained on millions of noisy image-text pairs, zero-shot CLIP can be directly applied to downstream classification task and achieve satisfying accuracy without any task-related training data (for example, 86.29\% on Caltech101 with ResNet-50 \cite{ResNet} visual encoder).

\subsection{Overfitting Issue in CoOp}
Formally, Context Optimization (CoOp) \cite{CoOp} first initializes a total of $M$ learnable word embeddings $\boldsymbol{v}=[\mathbf{v}_1,...,\mathbf{v}_M]\in \mathbb{R}^{d\times M}$.
Then, $\boldsymbol{t}_i=[\boldsymbol{v},\boldsymbol{c}_i]$ concatenating the learnable prompt $\boldsymbol{v}$ and the fixed category name embedding $\boldsymbol{c}_i$ is fed to the encoder $g(\cdot)$ and mapped to the text feature $\mathbf{w}_i$.
Under the few-shot setting, CoOp learns the embedding $\boldsymbol{v}$ on the labeled data $\{(\mathbf{x}_j,y_j)\}$ by minimizing the cross entropy loss:
\begin{equation}
\label{eq: coop}
L_{ce}(\boldsymbol{v})=-\sum_j \log p(y_j|\mathbf{x}_j).
\end{equation}
As reported in \cite{CoOp}, CoOp boosts the classification accuracy on the Caltech101 dataset \cite{caltech} from 86.29\% to 87.53\% under the 1-shot setting, with visual encoder being ResNet-50.
However, as reported in \cite{CoCoOp,zhu2022prompt,derakhshani2022variational} and our experimental results, two aspects of overfitting problem limit the effectiveness of CoOp on base classes and novel classes.
As shown in Fig. \ref{fig: base and novel}(a), the test accuracy on base categories first improves in the early training stage (from 1st to 10th epoch) and then worsens in the later training stage (from 11th to 50th epoch). Fig. \ref{fig: base and novel}(b) shows that the test accuracy on novel categories keeps declining.

Then, Conditional Context Optimization (CoCoOp) \cite{CoCoOp} is proposed to overcome the overfitting issue in CoOp using the image-specific prompt.
In particular, the prompt embedding $\boldsymbol{v}$ is added with a residual vector, which is generated by a learnable encoder fed with the image feature $\mathbf{z}$. 
However, CoCoOp cannot effectively eliminate overfitting in CoOp.

\subsection{Gradient Flow}
Gradient flow \cite{gradientflow,liu2017stein,arbel2019maximum,liutkus2019sliced} is a deterministic model of stochastic gradient descent (SGD), defined by the \textit{total gradient} differential equation:
\begin{equation}
\label{eq: gradient flow1}
\dot{\Phi}_t=\frac{\mathrm{d}\Phi}{\mathrm{d}t}=-g(\Phi),\quad g(\Phi)=E_{(x,y)\sim Train}\nabla_{\Phi}L(\Phi,x,y),
\end{equation}
where $\Phi(t)$ denotes the evolution of model parameters, and $g(\Phi)$ is the average gradient over the entire training set.
Consider the update $\Delta\Phi=-g(\Phi)\cdot\Delta t$, where $\Delta t$ represents the sum of learning rate over time. Thus, SGD can be viewed as a discretization of gradient flow.

\section{Methodology}
In this section, we first study the overfitting problem in prompt tuning by observing the gradient flows (Sec. \ref{sec: Analyzing gradient flow in prompt tuning}), 
and then conduct comparative experiments to identify the cause of the overfitting phenomenon (Sec. \ref{sec: Cause of overfitting phenomenon}). 
Finally, we propose Subspace Prompt Tuning (\textit{Sub}PT) to eliminate the overfitting problem (Sec. \ref{sec: Subspace Prompt Tuning}),
and introduce Novel Feature Learner (NFL) to enhance the generalization performance on novel categories (Sec. \ref{sec: Novel Feature Learner}). 

\subsection{Analyzing Gradient Flow in Prompt Tuning}
\label{sec: Analyzing gradient flow in prompt tuning}
The no free lunch theorem \cite{NoFreeLunch,GradientStarvation} states that it is impossible to learn without making proper assumptions on the distribution of training set and test set, so the i.i.d. assumption and empirical risk minimization have become the most popular in the machine learning community.
However, the i.i.d. assumption does not hold in many practical applications, leading to the learned parameters not being generalizable from the training set to the test set.
Therefore, an intuitive idea to understand the overfitting problem in prompt tuning is that CoOp favors the generalizable/spurious feature direction in the early/later training stage. 

\begin{figure}[t]
\centering
\scalebox{0.98}{
\subfloat[\footnotesize Early stage]{
\includegraphics[width=0.48\linewidth]{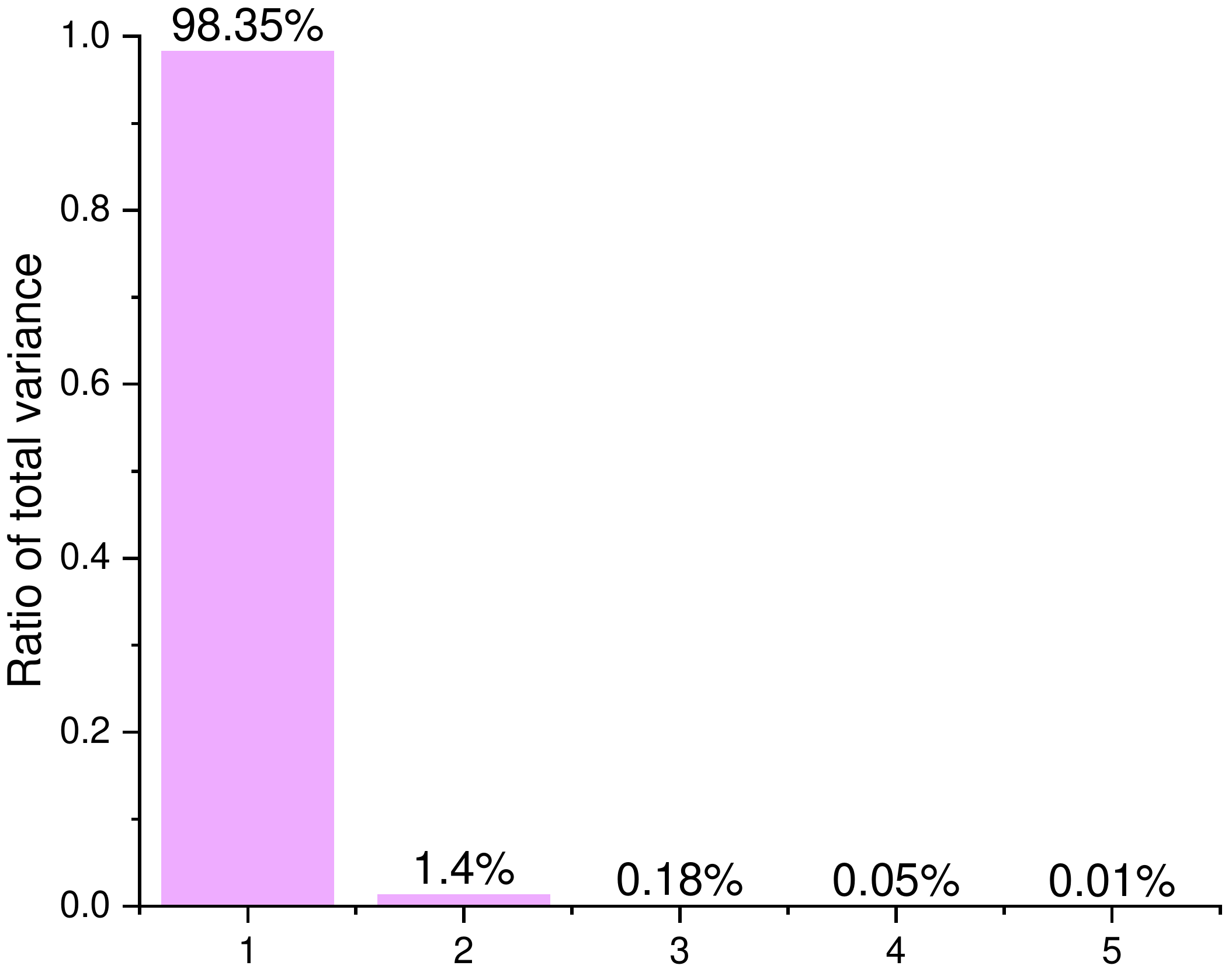}
\label{fig: base accuracy}
}
\hfill
\subfloat[\footnotesize Later stage]{
\includegraphics[width=0.48\linewidth]{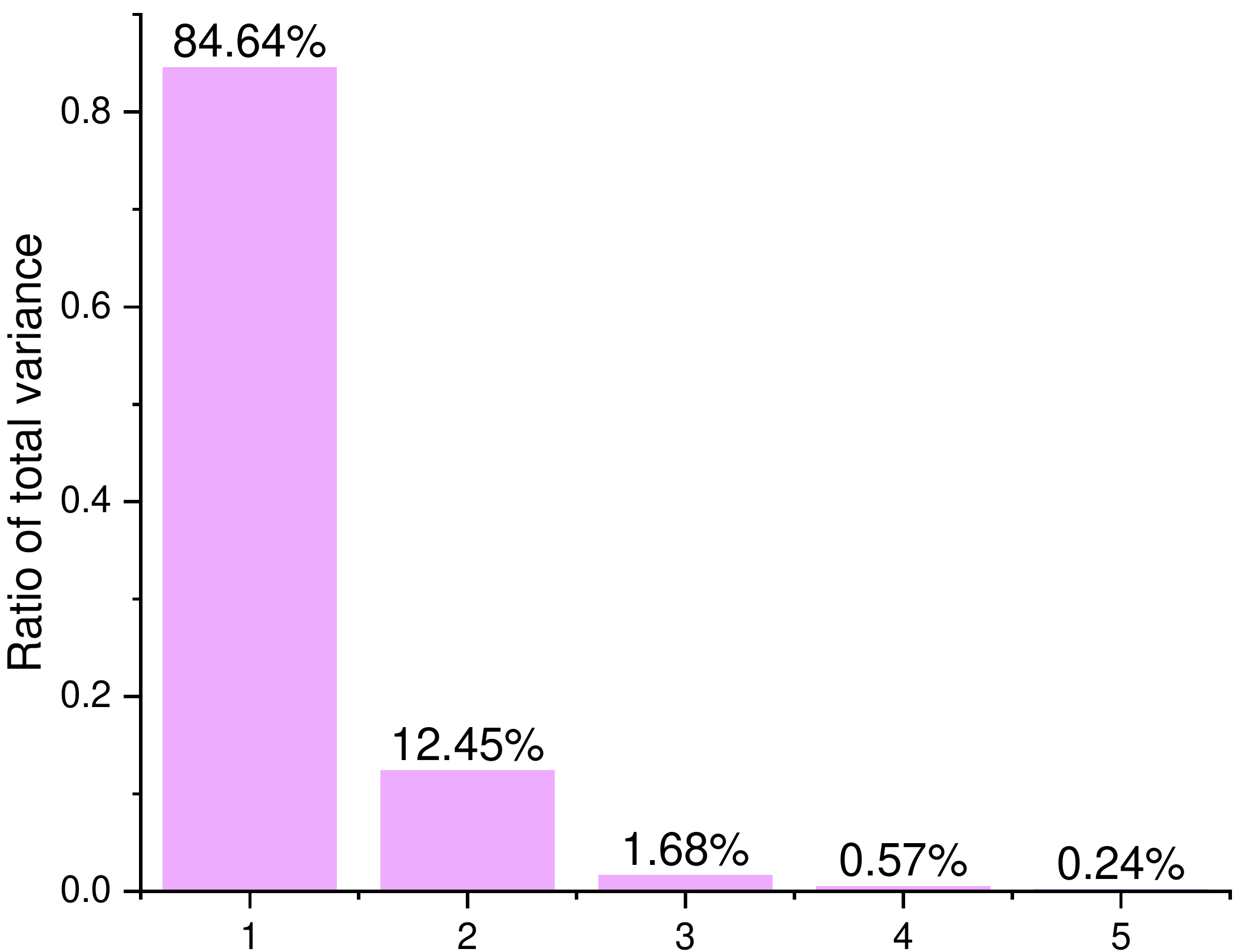}
\label{fig: novel accuracy}
}
}
\caption{\textbf{Variance distribution of PCA components}. The dataset is Caltech101. The spectrum of gradient flow is dominated by only a few components. (a) The early training stage (from 1st to 10th epoch). (b) The later training stage (from 31st to 50th epoch).}
\label{fig: PCA}
\end{figure}

To investigate the training direction, we formulate the gradient flow of the learnable prompt $\boldsymbol{v}$ as
\begin{align}
\label{eq: gradient flow2}
\dot{\boldsymbol{v}_t}
&=-\nabla L_{ce}(\boldsymbol{v}_t) \nonumber\\ 
&=-\nabla_{\boldsymbol{v}}g([\boldsymbol{v}_t,\mathcal{C}])^\intercal \cdot \nabla_{g([\boldsymbol{v}_t,\mathcal{C}])} L_{ce}(\mathcal{Z},g([\boldsymbol{v}_t,\mathcal{C}])),
\end{align}
where $\mathcal{C}$ and $\mathcal{Z}$ stand for the set of category names and image features appeared in the training process, respectively, and $L_{ce}$ refers to the softmax cross entropy loss.
The second term can be confirmed as a constant matrix\footnote{It is because that $\nabla_{\mathbf{z}^\intercal\mathbf{w}}L_{ce}=\mathbf{y}-\mathbf{\sigma (\mathbf{z}^\intercal\mathbf{w})}$ and $\nabla_{\mathbf{w}}\mathbf{z}^\intercal\mathbf{w}=\mathbf{z}$. ($\mathbf{y}$ denotes the labels in training set, $\sigma(\cdot)$ denote the softmax function, and $\mathbf{z}\in\mathcal{Z}$).} since both $\mathcal{C}$ and $\mathcal{Z}$ are fixed,
so the gradient flow $\dot{\boldsymbol{v}_t}$ is governed by the first term $\nabla_{\boldsymbol{v}}g([\boldsymbol{v}_t,\mathcal{C}])$, which denotes the gradient from output to input on the text encoder $g(\cdot)$ and varies over time $t$.
However, computing $\nabla_{\boldsymbol{v}}g([\boldsymbol{v}_t,\mathcal{C}])$ is still a laborious task.
Recalling the recent work \cite{DLDR}, Li \etal\; prove that 
both $\nabla_{\boldsymbol{v}}g([\boldsymbol{v}_t,\mathcal{C}])$ and $\dot{\boldsymbol{v}_t}$ can be approximated by low-rank matrices, 
because only a small part of eigenvalues dominates the spectrum of neural tangent kernel (NTK) under the infinite-width assumption \cite{NEURIPS2019_0d1a9651}, and NTK is defined as $\Theta_t=\nabla_{\boldsymbol{v}}g([\boldsymbol{v}_t,\mathcal{C}])\cdot\nabla_{\boldsymbol{v}}g([\boldsymbol{v}_t,\mathcal{C}])^\intercal$.
In practice, we follow \cite{DLDR} to first conduct an entire process of CoOp training and save the embedding checkpoint when every epoch ends. Then, we sample a trajectory of checkpoints $\boldsymbol{V}=\{\boldsymbol{v}_{t_1},...,\boldsymbol{v}_{t_2}\}\in\mathbb{R}^{(t_2-t_1)\times (d\times M)}$ and conduct principal component analysis (PCA) to compute the eigenvectors $\boldsymbol{U}=\{\boldsymbol{u}_{1},...,\boldsymbol{u}_{r}\}\in\mathbb{R}^{r \times (d\times M)}$ characterizing the gradient flow $\dot{\boldsymbol{v}_t}$, because the checkpoint trajectory can be viewed as a discretization of $\dot{\boldsymbol{v}_t}$ with some certain affine transformation. %
The PCA procedure is formulated as
\begin{equation}
\label{eq: pca}
\max_{\boldsymbol{U}}\| \boldsymbol{V} - \boldsymbol{V}\boldsymbol{U}^\intercal\boldsymbol{U} \|_F^2,\quad \text{s.t.} \;\; \boldsymbol{U}\boldsymbol{U}^\intercal=\boldsymbol I.
\end{equation}
Each checkpoint in $\boldsymbol{V}$ is centralized, and the eigenvectors in  $\boldsymbol{U}$ are sorted in descending order with respect to eigenvalues. 
We test with two trajectories within a total of 50 epochs: the early training stage (from 1st to 10th epoch) and the later training stage (from 31st to 50th epoch). As shown in Fig. \ref{fig: PCA}, the first eigenvector always occupies approximately 90\% of total variance, agreeing with the aforementioned analysis.
As a result, we can characterize the gradient flow of each training stage $\dot{\boldsymbol{v}_t}$ with its first eigenvector $\boldsymbol{u}_1$.

\begin{table}[t]
\renewcommand\arraystretch{1.32}
\begin{center}
\caption{Display on the inner product between two dominant eigenvectors across various datasets.}
\vspace{-1.5mm}
\label{tab: orthogonal}
\scalebox{.9}{
\begin{tabular}{l | c c c c c}
\hline
Dataset       & Caltech101 & Oxford Pets & Stanford Cars & DTD
\\ \hline
Inner product & -0.0099    & -0.0124     & -0.0547 & -0.0559
\\\hline 
\end{tabular}
}
\end{center}
\end{table}

Regarding our initial idea, we measure the inner product between the dominant eigenvectors of gradient flows in the early and later training stage across different datasets.
As shown in Table \ref{tab: orthogonal}, the inner product is always close to zero, meaning that the two gradient flows are almost orthogonal. 
Such experimental results reveal that CoOp does learn different features in early and later training stages.

\begin{figure*}[t]
    \centering
    \scalebox{0.53}{
    \includegraphics{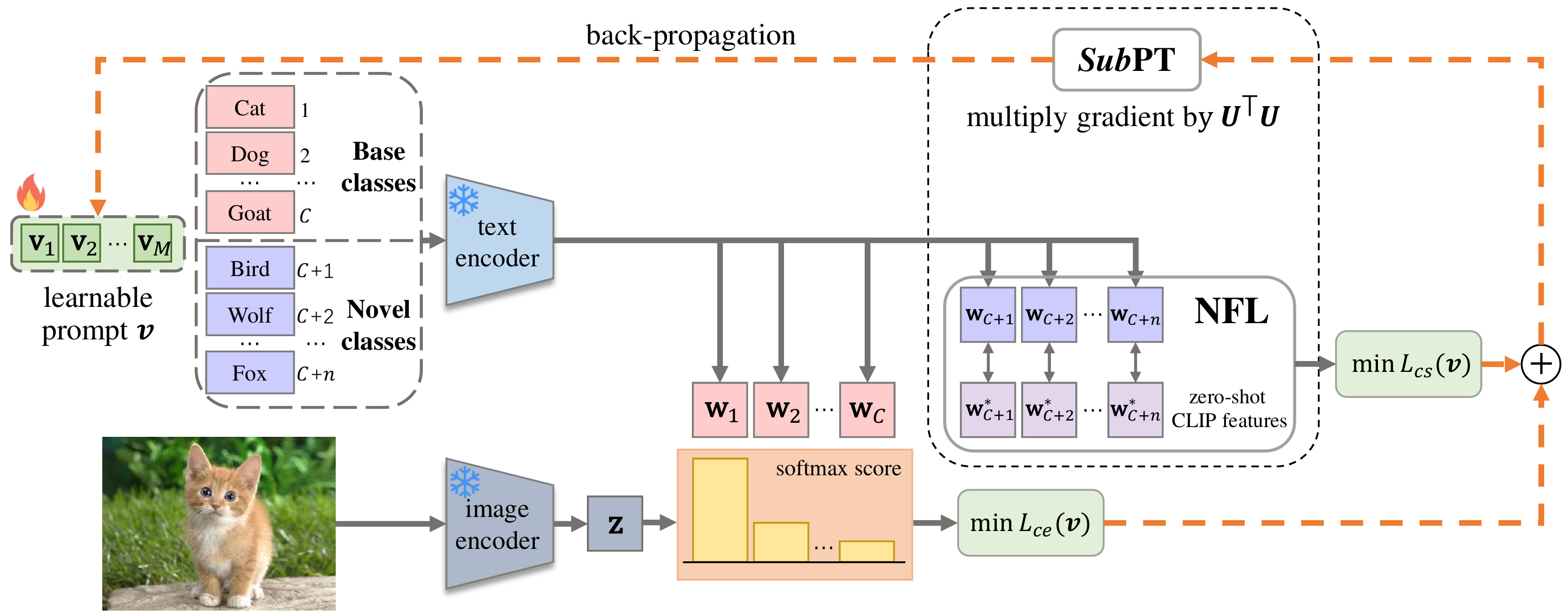}
    }
    \caption{\textbf{Overview of the proposed Subspace Prompt Tuning (\textit{Sub}PT) and Novel Feature Learner (NFL)}, surrounded by the black dotted box. To eliminate spurious components and mitigate overfitting, we project the gradient onto the low-rank \textbf{subspace} spanned by the dominant eigenvectors $\boldsymbol{U}$ of early-stage gradient flow during back-propagation. \textbf{NFL} learns text features towards zero-shot CLIP features on novel categories to enhance the generalization ability of the learned prompt embedding beyond the training set.}
    \label{fig: main}
\end{figure*}
\begin{figure}[t]
\centering
\vspace{-3mm}
\includegraphics[width=0.6\linewidth]{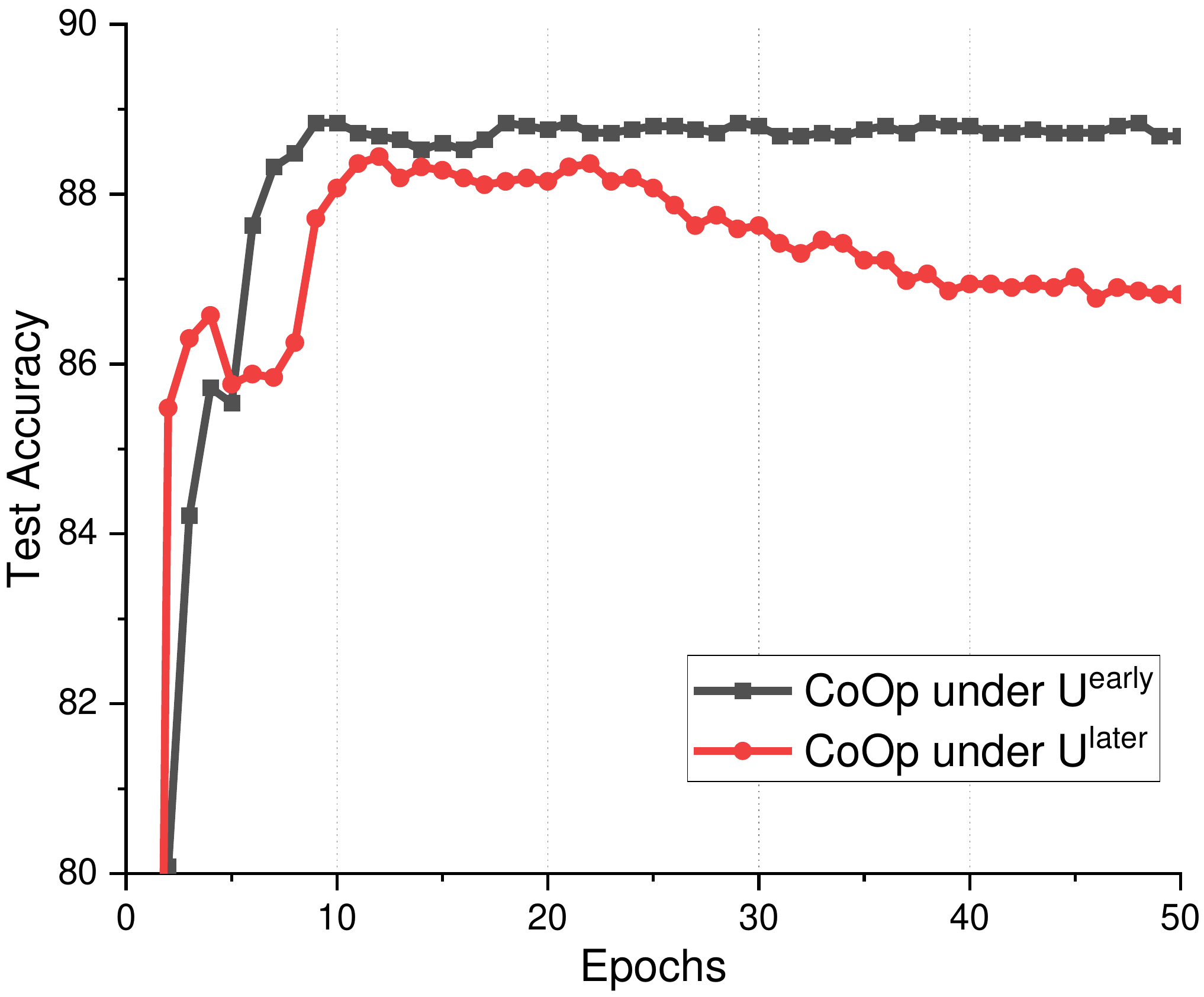}
\vspace{-2mm}
\caption{Comparison of the base class test set accuracy during the training process with $\boldsymbol{U}$ being $\boldsymbol{U}^{\text{early}}$ and $\boldsymbol{U}^{\text{later}}$, respectively. The dataset is Caltech101. 
The overfitting phenomenon occurs when $\boldsymbol{U}$ is $\boldsymbol{U}^{\text{later}}$, and can be mitigated using $\boldsymbol{U}^{\text{early}}$.}
\label{fig: early and later}
\end{figure}

\subsection{Cause of Overfitting}
\label{sec: Cause of overfitting phenomenon}
We design experiments to investigate whether CoOp favors generalizable/spurious features in the early/later training stage, leading to the non-overfitting/overfitting phenomenon.
For clarity, we first provide the experiment details and then provide an explanation.
In particular, we first conduct an entire process of CoOp training and save the embedding checkpoint at the end of each epoch.
On the basis of the observations presented in Sec. \ref{sec: Analyzing gradient flow in prompt tuning}, we compute two sets of eigenvectors ($\boldsymbol{U}^{\text{early}}$ and $\boldsymbol{U}^{\text{later}}$) from the checkpoints sampled in early and later training stages ($\{\boldsymbol{v}_t\}_{t=1}^{10}$ and $\{\boldsymbol{v}_t\}_{t=31}^{50}$) to represent the corresponding gradient flows, respectively.
Then, we rerun CoOp twice from the same initial embedding, and force two training processes towards the main components of these two gradient flows.
Similar to \cite{DLDR} and \cite{Larsen}, we multiply the gradient in each training step by the projection matrix $\boldsymbol{U}^\intercal\boldsymbol{U}\in\mathbb{R}^{(d\times M)\times (d\times M)}$, and the update formula of CoOp can be rewritten as
\begin{equation}
\label{eq: sub-coop}
\boldsymbol{v}\leftarrow\boldsymbol{v}-\alpha\cdot \boldsymbol{U}^\intercal\boldsymbol{U} \frac{\partial L_{ce}(\boldsymbol{v})}{\partial \boldsymbol{v}}.
\end{equation}
Note that we slightly abuse the notations here as both $\boldsymbol{v}$ and $\partial L_{ce}(\boldsymbol{v})/\partial \boldsymbol{v}$ are vectorized.
We omit the superscript of $\boldsymbol{U}$ without loss of generality.
$\alpha$ denotes the learning rate. 
In fact, the gradient vector $\partial L_{ce}(\boldsymbol{v})/\partial \boldsymbol{v}$ is projected onto the low-rank subspace spanned by eigenvectors. In this manner, the components in the directions orthogonal to eigenvectors $\boldsymbol{U}$ can be eliminated.
Thus, when we specify $\boldsymbol{U}$ in Eq. (\ref{eq: sub-coop}) as $\boldsymbol{U}^{\text{early}}$ or $\boldsymbol{U}^{\text{later}}$, the entire training process is forced towards the corresponding direction.

We take the Caltech101 \cite{caltech} dataset under 1-shot setting as an example. 
As shown in Fig. \ref{fig: early and later}, with $\boldsymbol{U}$ being $\boldsymbol{U}^{\text{early}}$, the test set performance increases to the best and never drops during the whole training process.
By contrast, with $\boldsymbol{U}$ being $\boldsymbol{U}^{\text{later}}$, the test set performance shares the similar trend with that of the original CoOp.  
Such experimental results indicates that CoOp does favor spurious features in the later training stage, leading to the overfitting phenomenon.

\begin{figure}[t]
    \centering
    \scalebox{0.45}{
    \includegraphics{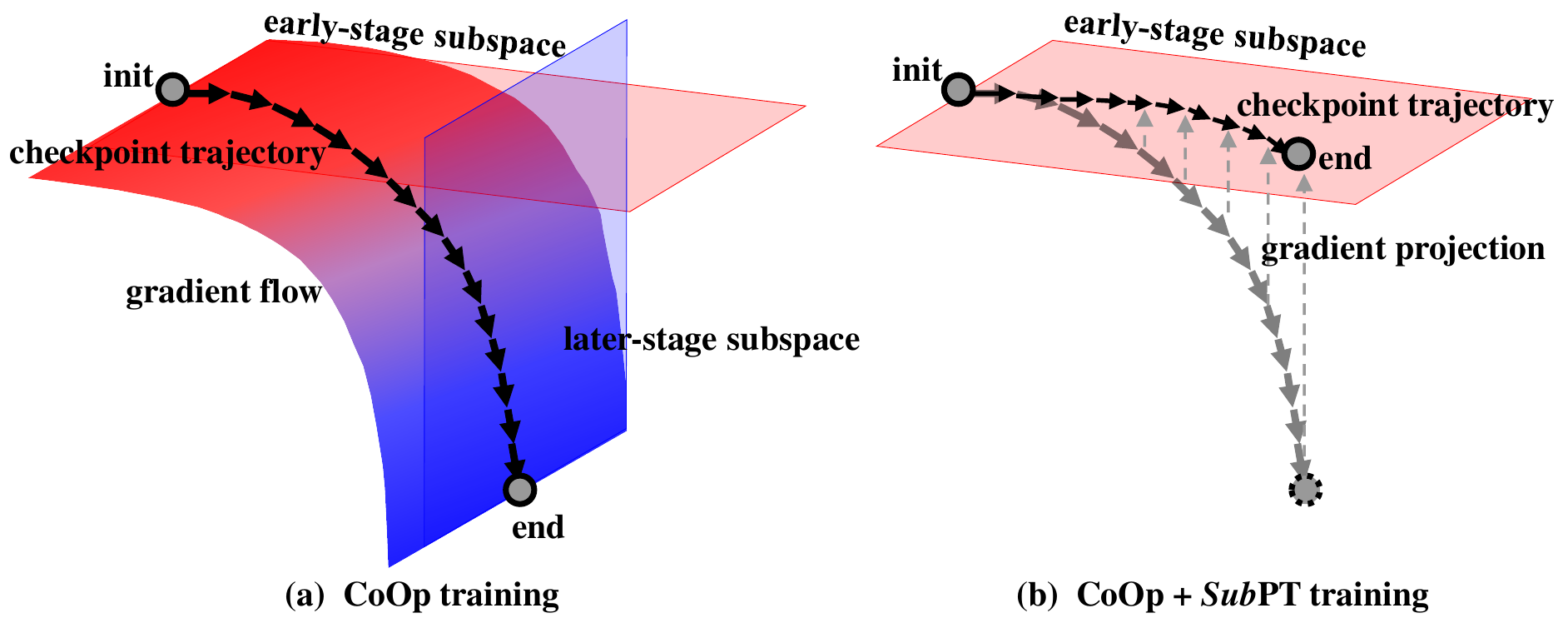}
    }
    \caption{An intuitive understanding on \textbf{Subspace Prompt Tuning (\textit{Sub}PT)}. We first construct the subspace spanned by prompt embedding checkpoints in early training stage, then we rerun CoOp from the same initial point and project gradients onto the constructed subspace in back-propagation to eliminate the spurious components. 
    The rationale behind \textit{Sub}PT \textcolor{blue}{is} that the spurious components are orthogonal to the early-stage subspace.}
    \label{fig: first-page}
\end{figure}

\noindent\textbf{Remark}.
We emphasize the fact about the above experiment that when specifying $\boldsymbol{U}$ as $\boldsymbol{U}^{\text{later}}$, the test performance first improves and then worsens, instead of keeping decreasing as the original CoOp does from 31st to 50th epoch.
This can be explained as the PCA procedure (refer to Eq. (\ref{eq: pca})) takes an unordered set of embedding checkpoints as inputs. Consequently, the computed eigenvectors $\boldsymbol{U}^{\text{later}}$ reflect the learned feature direction instead of the trajectory direction.

\subsection{Subspace Prompt Tuning (\textit{Sub}PT)}
\label{sec: Subspace Prompt Tuning}
On the basis of the aforementioned observations and analysis, it is then easy and straightforward to mitigate the overfitting issue in prompt tuning in three steps. 
\begin{enumerate}[Step 1:]
\item Conduct prompt tuning and save the embedding checkpoint $\boldsymbol{v}_t$ at the end of every epoch.
\item Compute dominant eigenvectors \small $\boldsymbol{U}^{\text{early}}=\{\boldsymbol{u}_{1},\dots,\boldsymbol{u}_{r}\}$ \normalsize of the saved checkpoints $\boldsymbol{V}_{\text{early}}=\{\boldsymbol{v}_1,...,\boldsymbol{v}_{t_{\text{early}}}\}$ in the specified early training stage (refer to Eq. (\ref{eq: pca})). $t_{\text{early}}$ denotes the last epoch of the early training stage. $r$ represents the number of eigenvectors ($r\leqslant t_{\text{early}}$).
\item Rerun prompt tuning from the same initial point, and project the gradients in back-propagation onto the low-rank subspace spanned by $\boldsymbol{U}^{\text{early}}$ during the entire training process (refer to Eq. (\ref{eq: sub-coop})).
\end{enumerate}
We name this approach Subspace Prompt Tuning (\textit{Sub}PT),
in which the generalizable/spurious components in gradients are maintained/eliminated during the multiplication process (see Figs.~\ref{fig: main} and \ref{fig: first-page} for the overview of proposed method and the intuitive understanding). 
Figs.~\ref{fig: base and novel}(a) and \ref{fig: base and novel}(b) demonstrate that \textit{Sub}PT can address the overfitting issue effectively.

\subsection{Novel Feature Learner (NFL)}
\label{sec: Novel Feature Learner}
As CoOp only optimizes for the base categories within the training set, CoOp inevitably hurts the zero-shot capability towards novel categories.
Thus, zero-shot CLIP can still be a good teacher to regularize CoOp training.
In this section, we present the Novel Feature Learner (NFL) to enhance the generalization ability, with no need for any image training data.
Concretely, NFL encourages the text feature similarity on novel categories between CoOp and zero-shot CLIP.
As shown in Fig. {\ref{fig: main}}, NFL contains three components:
text features on novel categories generated from learnable prompts, zero-shot CLIP features in correspondence, and a feature similarity loss function $L_{cs}$.
We first encode the learnable prompt embedding combining novel category name embeddings into text features $\{\mathbf{w}_i\}_{i=C+1}^{C+n}$ with the text encoder $g(\cdot)$. Then, we utilize zero-shot CLIP to obtain corresponding text features $\{\mathbf{w}_i^\ast\}_{i=C+1}^{C+n}$ with the handcrafted prompts, and further optimize the feature similarities between them with $L_{cs}$.
We give more details in the following part.

\vspace{1mm}
\noindent\textbf{Novel category selection.}
We consider the following two scenes.
If the novel category names are unknown during training (such as the few-shot classification), then we randomly select $n$ categories from 1000 ImageNet \cite{imagenet} categories, disjoint from the base categories within the training set.
In the experiments, as the performance gain brought by NFL remains similar as $n$ increases from 100 to 1000, we fix $n$ as 100 for computational convenience.
Moreover, if the novel category names are known during training (such as the base-to-novel classification), then we directly utilize these novel category names for NFL.

\vspace{1mm}
\noindent\textbf{Zero-shot CLIP features.}
To obtain corresponding zero-shot CLIP features, we select 80 handcrafted prompt templates provided by \cite{CLIP}, such as ``a good photo of the [CLASS]" and ``an origami [CLASS]", and further compute the averaged text feature for each category.

\vspace{1mm}
\noindent\textbf{Feature similarity loss.}
As the zero-shot CLIP has great generalization capability, we propose to minimize the feature similarities between $\{\mathbf{w}_i\}_{i=C+1}^{C+n}$ and $\{\mathbf{w}_i^\ast\}_{i=C+1}^{C+n}$ to enhance the performance of learned prompt embeddings towards novel categories.
We choose the cosine similarity as the feature loss function, and the NFL is formulated as
\begin{equation}
\label{eq: NFL}
L_{cs}(\boldsymbol{v})=\frac{1}{n}\sum_{i=C+1}^{C+n}(1-\mathbf{w}_i^\intercal \mathbf{w}_i^\ast).
\end{equation}
As NFL does not require any image training data, we can conveniently apply it as an extra loss term, and the total loss is written as
\begin{equation}
\label{eq: total}
L(\boldsymbol{v})=L_{ce}(\boldsymbol{v})+\lambda \cdot L_{cs}(\boldsymbol{v}).
\end{equation}

\section{Experiments}
\label{sec:exp}
In this section, we evaluate our proposed \textit{Sub}PT and NFL on three downstream vision tasks: image classification, open-vocabulary object detection, and zero-shot semantic segmentation. 
We describe the datasets and evaluation metrics in detail, along with baselines and implementation details.
We also conduct ablation studies on all the hyper-parameters in the proposed approach.

\subsection{Image Classification}
We consider three problem settings for the image classification task: \textit{few-shot classification}, \textit{base-to-novel generalization}, and \textit{domain generalization}.

\vspace{1mm}
\noindent\textbf{Datasets.} 
For few-shot classification and base-to-novel generalization, we follow CoOp \cite{CoOp} and CoCoOp \cite{CoCoOp} to rely on 11 classification datasets, including generic object classification datasets (\ie \;ImageNet \cite{imagenet} and Caltech101 \cite{caltech}), fine-grained visual classification (\ie \;Oxford Pets \cite{oxfordpets}, Stanford Cars \cite{stanfordcars}, Flowers 102 \cite{oxfordflowers}, Food 101 \cite{food101} and FGVC Aircraft \cite{fgvcaircraft}), scene recognition (\ie \;SUN 397 \cite{sun397}), texture classification (\ie \;DTD \cite{dtd}), satellite imagery recognition (\ie \;EuroSAT \cite{eurosat}), and action recognition (\ie \;UCF 101 \cite{ucf101}). 
For base-to-novel generalization, we split the first half of categories as base classes and the second half as novel classes within each dataset.
For domain generalization, we choose ImageNet as source domain dataset and evaluate performance on four target domain datasets: ImageNet-V2 \cite{imagenetv2}, ImageNet-Sketch \cite{imagenetsketch}, ImageNet-A \cite{imagenet-a}, and ImageNet-R \cite{imagenet-r}.

\vspace{1mm}
\noindent\textbf{Evaluation Metrics.}
For all problem settings, we report the average accuracy over three different random seeds. We also report the harmonic mean HM=2$\times$(base$\times$new)/(base+new) for base-to-novel generalization.

\vspace{1mm}
\noindent\textbf{Baselines.}
We compare our proposed approach with both zero-shot CLIP \cite{CLIP} (based on the handcrafted prompt) and CoOp \cite{CoOp}.
As CoCoOp \cite{CoCoOp} can degrade the base accuracy of CoOp as reported in \cite{CoCoOp} and is only designed to improve the generalization ability of learned prompt, we only compare with CoCoOp on the base-to-novel generalization, cross-data transfer, and domain generalization settings.

\vspace{1mm}
\noindent\textbf{Implementation Details.}
We follow the official implementation of CoOp\footnote{\url{https://github.com/KaiyangZhou/CoOp/blob/main/COOP.md}} \cite{CoOp} and CoCoOp\footnote{\url{https://github.com/KaiyangZhou/CoOp/blob/main/COCOOP.md}} \cite{CoCoOp} to adopt few-shot learning for all problem settings.
For few-shot classification, we conduct prompt tuning with 1, 2, 4, 8, and 16 shots respectively and evaluate performance on the entire test set.
For base-to-novel generalization, 
cross-data transfer,
and domain generalization, we adopt the 4-shots training. 
The prompt length $M$ is fixed as 16 and the prompt embedding is initialized with Gaussian distribution $\mathcal N(0, 0.02)$.
We follow the same training schedule and data augmentation strategies in CoOp \cite{CoOp}.
The visual encoder is specified as ResNet-50 for all experiments.

The only two hyper-parameters in \textit{Sub}PT are $t_{\text{early}}$ and $r$ (we will study the dependency on them in the ablations later).
Here, we set ($t_{\text{early}}$, $r$) as (10, 5) for 1 shot, (20, 10) for 2 shots, (30, 10) for 4 shots, (40, 10) for 8 shots, and (50, 10) for 16 shots. 
We observe in the experiments that the overfitting problem does not occur on the Flowers 102 and EuroSAT datasets, so we set $t_{\text{early}}$ as the total epoch for these two datasets. 
For NFL, we set the loss weight $\lambda$ as 1.0.

\begin{figure*}[t]
    \centering
    \scalebox{0.71}{
    \includegraphics{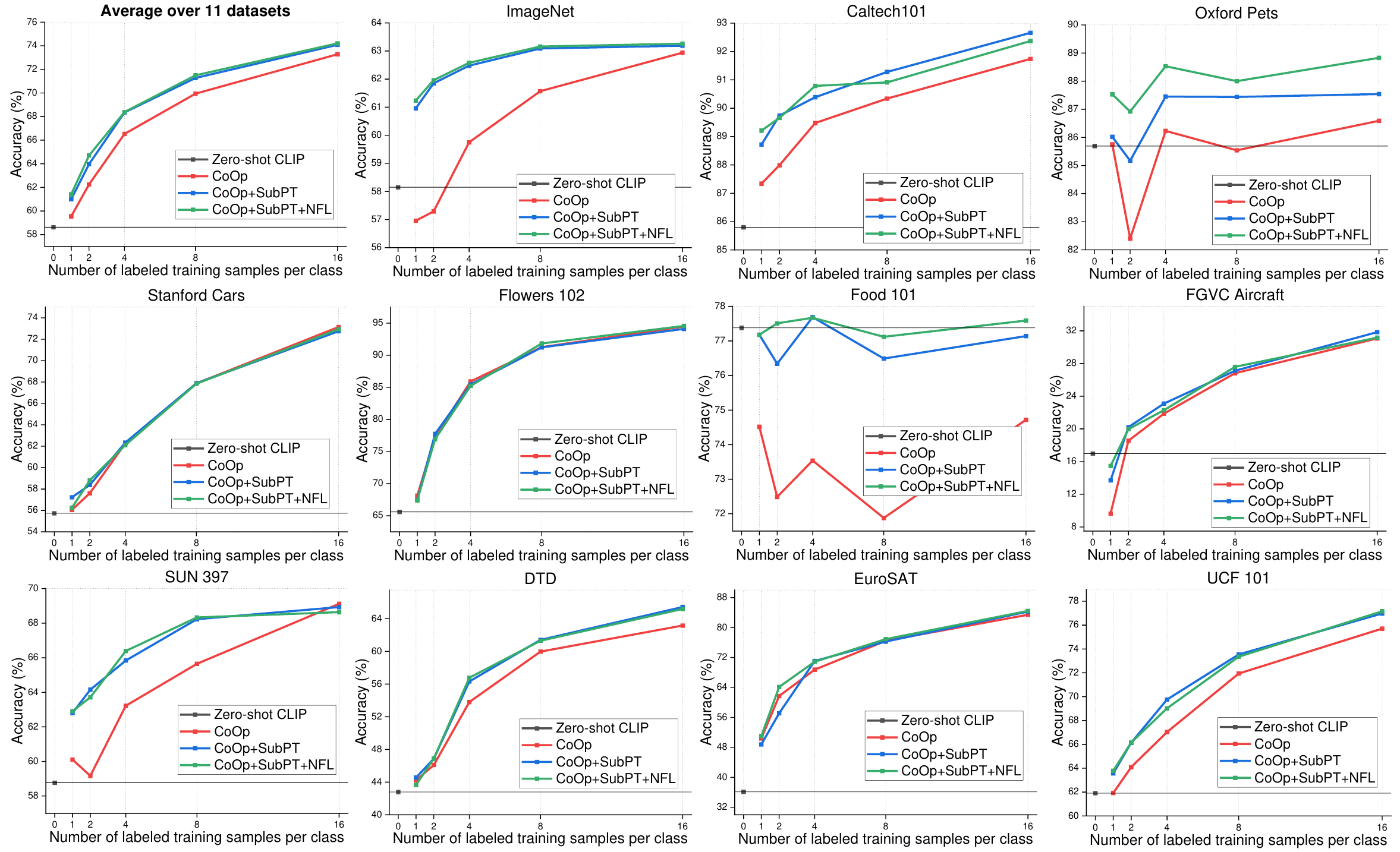}
    }
    \vspace{-1mm}
    \caption{\footnotesize{\textbf{Comparison with zero-shot CLIP and CoOp on few-shot classification of 11 datasets}. Our proposed approaches ``CoOp+SubPT" and ``CoOp+SubPT+NFL" can consistently improve the performances of CoOp.}}
    \label{fig: few-shot}
\end{figure*}

\vspace{1mm}
\noindent\textbf{Few-Shot Classification Results.}
Fig. \ref{fig: few-shot} shows the comparison with baseline approaches on 11 classification datasets. 
The average accuracies over all datasets are also displayed in the first subfigure. 
Compared with the zero-shot CLIP baseline, the proposed \textit{Sub}PT can consistently and substantially outperform the average accuracy by 2.37\% with 1 training sample per class and 15.45\% with 16 training samples per class.
Then, NFL can further increase the accuracy by 0.11\% to 0.73\%.
In addition, \textit{Sub}PT plus NFL can also boost the average performance of CoOp with a 3.14\% relative gain in the 1-shot scenario and a 1.23\% relative gain in the 16-shot scenario.
For example, on Food 101 dataset, the performance of CoOp+\textit{Sub}PT+NFL can surpass that of CoOp by 3.57\% in the 1-shot scenario and 3.84\% in the 16-shot scenario.
These results validates the effectiveness of both \textit{Sub}PT and NFL in mitigating the overfitting issue in CoOp.

\begin{figure*}[t]
    \centering
    \scalebox{0.53}{
    \includegraphics{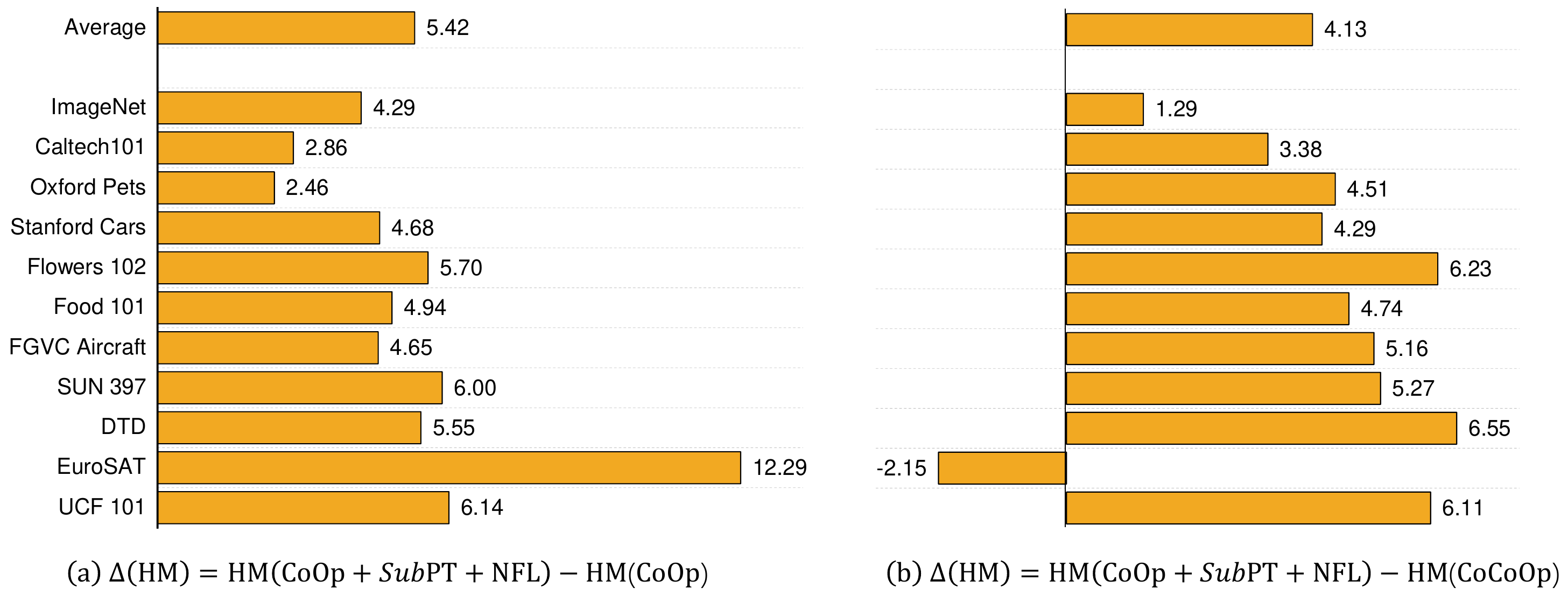}
    }
    \vspace{-2mm}
    \caption{\textbf{Absolute performance improvement of the proposed approach over CoOp and CoCoOp} in terms of harmonic mean accuracy over 11 classification datasets. \textit{Sub}PT and NFL can always enhance the performance of CoOp and surpass that of CoCoOp in 10 out of 11 datasets.}
    \label{fig: base2novel}
\end{figure*}

\vspace{1mm}
\noindent\textbf{Base-to-Novel Generalization Results.}
Table \ref{tab: base2novel} presents the performance of \textit{Sub}PT and NFL in the base-to-novel generalization setting on 11 classification datasets. 
The overfitting problem in CoOp is reflected by the divergence of test set accuracies between base and novel classes. For example, CoOp can boost the average performance of zero-shot CLIP on base classes with an absolute 6.93\% margin over 11 datasets, but can also degrade CLIP's performance on novel classes by an absolute 10.30\% margin. 
CoCoOp can remedy the CoOp's novel class accuracy with 1.67\% by changing the static prompts into dynamic ones, but the remedied accuracy is still far below that of zero-shot CLIP (60.42\% \textit{vs} 69.05\%).
By contrast, adding \textit{Sub}PT to CoOp mitigates the overfitting problem on both base and novel classes, as the base accuracy improves from 72.10\% to 72.57\% and novel accuracy improves from 58.75\% to 62.30\%.
Furthermore, NFL can further raise the novel accuracy up to 66.91\%, illustrating the effectiveness of NFL.
In detail, Fig. \ref{fig: base2novel} depicts the absolute performance improvement of \textit{Sub}PT and NFL over CoOp and CoCoOp in terms of the harmonic mean accuracy. 
Our proposed approach can consistently improve CoOp over 11 datasets and surpass CoCoOp expect for the EuroSAT dataset.

\begin{table*}[t]
\centering
\renewcommand\arraystretch{1.25}
\caption{Comparison with state-of-the-art approaches on base-to-novel generalization. HM stands for Harmonic Mean. \textbf{Best} and \textbf{\underline{second best}} results are highlighted.}
\vspace{-1.mm}
\scalebox{1.0}{
\begin{tabular}{l | c c c | c c c | c c c}
\hline
       & \multicolumn{3}{c|}{\textbf{\textit{Average}}} & \multicolumn{3}{c|}{ImageNet} & \multicolumn{3}{c}{Caltech101} \\
       & Base  & Novel & HM    & Base  & Novel & HM    & Base  & Novel & HM    \\\hline
Zero-shot CLIP   & 65.17 & \textbf{69.05} & \textbf{\underline{66.94}} & 64.39 & 60.05 & 62.14 & 90.77 & \textbf{\underline{91.05}} & 90.91 \\
CoOp   & 72.10 & 58.75 & 63.90 & 64.25 & 54.99 & 59.26 & 94.00 & 87.37 & 90.56 \\
CoCoOp & 71.91 & 60.42 & 65.19 & \textbf{\underline{66.49}} & 58.54 & 62.26 & \textbf{\underline{94.36}} & 86.10 & 90.04 \\
\rowcolor{black!10} CoOp + \textit{Sub}PT
       & \textbf{\underline{72.57}} & 62.30 & 66.37 & \textbf{66.91} & \textbf{\underline{60.32}} & \textbf{\underline{63.44}} & 94.35 & 89.27 & \textbf{\underline{91.74}} \\
\rowcolor{black!10} CoOp + \textit{Sub}PT + NFL
       & \textbf{72.96} & \textbf{\underline{66.91}} & \textbf{69.32} & 66.14 & \textbf{61.16} & \textbf{63.55} & \textbf{94.77} & \textbf{92.10} & \textbf{93.42} \\\hline
       & \multicolumn{3}{c|}{Oxford Pets} & \multicolumn{3}{c|}{Stanford Cars} & \multicolumn{3}{c}{Flowers 102} \\
       & Base  & Novel & HM    & Base  & Novel & HM    & Base  & Novel & HM    \\\hline
Zero-shot CLIP   & 90.06 & \textbf{\underline{94.18}} & 92.07 & 55.32 & \textbf{66.65} & 60.46 & 69.14 & \textbf{73.97} & 71.47 \\
CoOp   & 89.67 & 92.34 & 90.99 & \textbf{\underline{62.57}} & 55.24 & 58.68 & 86.80 & 63.55 & 73.38 \\
CoCoOp & 87.75 & 90.16 & 88.94 & 61.32 & 56.97 & 59.07 & 87.08 & 62.62 & 72.85 \\
\rowcolor{black!10} CoOp + \textit{Sub}PT
       & \textbf{\underline{90.96}} & 92.60 & 91.77 & \textbf{62.67} & 58.83 & \textbf{\underline{60.69}} & \textbf{88.75} & 63.22 & \textbf{\underline{73.84}} \\
\rowcolor{black!10} CoOp + \textit{Sub}PT + NFL
       & \textbf{91.65} & \textbf{\underline{95.32}} & 93.45 & 61.08 & \textbf{\underline{65.82}} & \textbf{63.36} & \textbf{\underline{88.19}} & \textbf{\underline{71.68}} & \textbf{79.08} \\\hline
       & \multicolumn{3}{c|}{Food 101} & \multicolumn{3}{c|}{FGVC Aircraft} & \multicolumn{3}{c}{SUN 397} \\
       & Base  & Novel & HM    & Base  & Novel & HM    & Base  & Novel & HM    \\\hline
Zero-shot CLIP   & \textbf{83.59} & \textbf{85.01} & \textbf{84.29} & 18.79 & \textbf{25.97} & \textbf{\underline{21.80}} & 66.78 & \textbf{\underline{70.51}} & 68.59 \\
CoOp   & 78.59 & 77.86 & 78.22 & \textbf{\underline{22.63}} & 16.92 & 19.36 & 71.16 & 62.20 & 66.38 \\
CoCoOp & 78.53 & 78.31 & 78.42 & 21.67 & 16.68 & 18.85 & 70.18 & 64.29 & 67.11 \\
\rowcolor{black!10} CoOp + \textit{Sub}PT 
       & 80.76 & 80.85 & 80.80 & 19.09 & 21.78 & 20.35 & \textbf{\underline{72.37}} & 68.14 & \textbf{\underline{70.19}} \\
\rowcolor{black!10} CoOp + \textit{Sub}PT + NFL
       & \textbf{\underline{82.20}} & \textbf{\underline{84.14}} & \textbf{\underline{83.16}} & \textbf{22.99} & \textbf{\underline{25.13}} & \textbf{24.01} & \textbf{73.35} & \textbf{71.44} & \textbf{72.38} \\\hline
       & \multicolumn{3}{c|}{DTD} & \multicolumn{3}{c|}{EuroSAT} & \multicolumn{3}{c}{UCF 101} \\
       & Base  & Novel & HM    & Base  & Novel & HM    & Base  & Novel & HM    \\\hline
Zero-shot CLIP   & 54.17 & \textbf{56.16} & \textbf{\underline{55.15}} & 54.83 & \textbf{66.18} & 59.97 & 69.03 & \textbf{69.82} & \textbf{\underline{69.42}} \\
CoOp   & \textbf{\underline{66.09}} & 42.75 & 51.92 & 85.10 & 35.67 & 50.27 & 72.23 & 57.33 & 63.92 \\
CoCoOp & 65.74 & 41.55 & 50.92 & \textbf{85.74} & \textbf{\underline{51.97}} & \textbf{64.71} & 72.13 & 57.44 & 63.95 \\
\rowcolor{black!10} CoOp + \textit{Sub}PT
       & \textbf{66.28} & 46.50 & 54.66 & 81.98 & 40.79 & 54.48 & \textbf{74.10} & 63.01 & 68.11 \\
\rowcolor{black!10} CoOp + \textit{Sub}PT + NFL   
       & 63.73 & \textbf{\underline{52.33}} & \textbf{57.47} & \textbf{\underline{85.71}} & 49.26 & \textbf{\underline{62.56}} & \textbf{\underline{72.71}} & \textbf{\underline{67.59}} & \textbf{70.06}
\\\hline
\end{tabular}
}
\label{tab: base2novel}
\end{table*}

\vspace{1mm}
\noindent\textbf{Cross-Dataset Transfer Results.}\;\;\;\; 
We compare our method with CoOp and CoCoOp by transferring prompt embeddings learned from ImageNet to 10 other datasets. As shown in Table {\ref{tab: cross-dataset}}, on the source dataset, CoCoOp outperforms CoOp by a 1.27\% margin, whereas CoOp equipped with \textit{Sub}PT can surpass CoCoOp by a margin of 1.47\%. In addition, NFL can further boost the classification accuracy to 62.58\% and achieve the best performances, indicating the effectiveness of our method on base classes. The performances show the similar trends on 10 target datasets, as our CoOp+\textit{Sub}PT+NFL can always beat CoCoOp (\eg, ours at 61.72\% \textit{vs} CoCoOp at 57.96\% on the UCF 101 dataset), which suggests the impressive generalization ability onto novel classes.

\vspace{1mm}
\noindent\textbf{Domain Generalization Results.}\;\;\;\;
As discussed in Sec. \ref{sec: Analyzing gradient flow in prompt tuning}, CoOp learns both generalizable and spurious features during the entire training process, whereas the latter can degrade the generalization ability of learned prompt embeddings onto the out-of-distribution and adversarial datasets.
As shown in Table \ref{tab: out-of-distribution}, CoOp's test set accuracies on four target domain datasets are always below that of the original ImageNet, although all of these datasets share the same categories. 
Similar to the base-to-novel generalization setting, CoCoOp can still enhance the generalization ability of CoOp but with insignificant improvements.
In comparison, adding \textit{Sub}PT to CoOp can persistently increase the performance of CoOp and even outperform that of zero-shot CLIP over four target datasets (for example, 57.58\% \textit{vs} 56.00\% on ImageNet-R), which verifies that our proposed \textit{Sub}PT can eliminate the spurious features during training.
In addition, adding NFL to CoOp plus \textit{Sub}PT further boosts the performances, confirming that as a regularization term, NFL can enhance the domain generalization ability of the learned prompt embeddings (\ie we set $t_{\text{early}}$ to 10 due to a more severe overfitting issue on ImageNet).

\subsection{Open-Vocabulary Object Detection}
Here, we consider the more challenging downstream vision task, that is \textit{open-vocabulary object detection (OVOD)}.

\vspace{1mm}
\noindent\textbf{Datasets and Evaluation Metrics.} \;\;
We follow the state-of-the-art work PromptDet \cite{PromptDet} to conduct experiments on the LVIS dataset \cite{LVIS}, which is annotated with bounding boxes and divided into 866 base categories and 337 novel categories.
For evaluation on LVIS v1.0 minival set, we report the mask Average Precision for novel categories AP$_{novel}$, in which a higher score indicates a better generalization onto novel category objects beyond the training set.

\vspace{1mm}
\noindent\textbf{Baselines.}
In terms of the prompts utilized in OVOD, we select both the zero-shot CLIP baseline with manual prompt ``a photo of [CLASS]" and the Regional Prompt Learning (RPL) \cite{PromptDet} as baselines, where RPL learns continuous prompt embedding on the cropped object patches within base categories in the CoOp manner. 
To examine the ability of \textit{Sub}PT against over-fitting problem, we equip RPL with \textit{Sub}PT to learn better prompt embeddings for OVOD.

\vspace{1mm}
\noindent\textbf{Implementation Details.}
Similar with the RPL process in PromptDet \cite{PromptDet}, we choose ViT-B/32 as the visual encoder backbone and conduct 1-shot training.
We randomly crop the input image within the range (40\%, 100\%), and follow the same training settings in CoOp \cite{CoOp}, with batch size being 32 and base learning rate being 2e-3 (adjusted by cosine annealing schedule).
After RPL, we conduct object detector training over the Mask-RCNN \cite{mask-rcnn} architecture with a ResNet-50-FPN backbone and train 12 epochs for all baselines, based on the official implementation of PromptDet\footnote{\url{https://github.com/fcjian/PromptDet}}.
Here, we omit the self-training procedure in PromptDet \cite{PromptDet} for fair comparisons.

\begin{table*}[t]
\centering
\renewcommand\arraystretch{1.25}
\caption{Comparison with state-of-the-art approaches on cross-dataset transfer. \textbf{Best} and {\textbf{\underline{second best}}} results are highlighted.}
\scalebox{1.02}{
\begin{tabular}{l | c | c c c c c c c c c c c}
\hline
& \textbf{Source}   & \multicolumn{11}{c}{\textbf{Target}} \\
& \rotatebox{90}{ImageNet}          & \rotatebox{90}{Caltech101} & \rotatebox{90}{Oxford Pets} & \rotatebox{90}{Stanford Cars} & \rotatebox{90}{Flowers 102} & \rotatebox{90}{Food 101} & \rotatebox{90}{FGVC Aircraft} & \rotatebox{90}{SUN 397} & \rotatebox{90}{DTD} & \rotatebox{90}{EuroSAT} & \rotatebox{90}{UCF 101} & \rotatebox{90}{\textit{\textbf{Average}}}
\\\hline
CoOp        & 59.74 & 84.89 & 81.32 & 53.50 & 56.88 & 73.59 & 13.43 & 54.83 & 32.05 & 25.52 & 54.72 & 53.68 \\
CoCoOp      & 61.01 & 87.32 & 86.42 & 53.47 & 62.50 & 76.52 & 14.28 & 57.91 & 37.31 & 20.66 & 57.96 & 55.93 \\
\rowcolor{black!10} CoOp + \textit{Sub}PT
            & \textbf{\underline{62.48}} & \textbf{88.37} & \textbf{\underline{87.14}} & \textbf{\underline{55.59}} & \textbf{64.26} & \textbf{\underline{77.37}} & \textbf{\underline{16.17}} & \textbf{\underline{60.85}} & \textbf{\underline{38.34}} & \textbf{\underline{28.74}} & \textbf{\underline{60.01}} & \textbf{\underline{58.12}} \\
\rowcolor{black!10} CoOp + \textit{Sub}PT + NFL
            & \textbf{62.58} & \textbf{\underline{88.28}} & \textbf{87.35} & \textbf{56.21}	& \textbf{\underline{63.38}}	& \textbf{77.83}	& \textbf{16.68}	& \textbf{61.81}	& \textbf{39.66}	& \textbf{29.06}	& \textbf{61.72} & \textbf{58.58} \\\hline
\end{tabular}
}
\label{tab: cross-dataset}
\end{table*}
\begin{table*}[t]
\centering
\renewcommand\arraystretch{1.25}
\caption{Comparison with state-of-the-art approaches on domain generalization. \textbf{Best} and \textbf{\underline{second best}} results are highlighted.}
\scalebox{1.02}{
\begin{tabular}{l | c | c c c c}
\hline
            & \textbf{Source}   & \multicolumn{4}{c}{\textbf{Target}} \\
            & ImageNet          & ImageNet-V2 & ImageNet-Sketch & ImageNet-A & ImageNet-R
\\\hline
Zero-shot CLIP        & 58.18             & 51.34       & 33.32           & 21.65      & 56.00 \\
CoOp        & 59.74             & 52.52       & 31.42           & 22.25      & 53.51 \\
CoCoOp      & 61.01             & 53.87       & 32.65           & 22.56      & 54.31 \\
\rowcolor{black!10} CoOp + \textit{Sub}PT
            & \textbf{\underline{62.48}}    & \textbf{\underline{55.36}}       & \textbf{\underline{34.73}}           & \textbf{\underline{23.95}}      & \textbf{\underline{57.58}} \\
\rowcolor{black!10} CoOp + \textit{Sub}PT + NFL
            & \textbf{62.58}           & \textbf{55.44}       & \textbf{35.30}             & \textbf{24.52}      & \textbf{58.64}
\\\hline
\end{tabular}
}
\label{tab: out-of-distribution}
\end{table*}
\begin{table}[t]
\centering
\renewcommand\arraystretch{1.25}
\caption{Performance on open-vocabulary object detection. \textbf{Best} results are highlighted.}
\vspace{-1.mm}
\scalebox{1.02}{
\begin{tabular}{c c c c c}
\hline
Model                     & RPL        & \textit{Sub}PT  & shots  & AP$_{novel}$ 
\\\hline
\!Baseline (manual prompt)\!  &            &           &   0    & 7.4 \\
\!PromptDet\_R\_50\_FPN\_1x\! & \checkmark &           &   1    & 6.4  \\
\rowcolor{black!10} \!PromptDet\_R\_50\_FPN\_1x\! & \checkmark & \checkmark&   1    & \textbf{9.7 (+3.3)} \\
\hline
\end{tabular}
}
\vspace{-.28mm}
\label{tab: owod}
\end{table}
\begin{table}[t]
\centering
\renewcommand\arraystretch{1.25}
\caption{Performance on zero-shot semantic segmentation. \textbf{Best} results are highlighted.}
\vspace{-1.mm}
\scalebox{1.02}{
\begin{tabular}{l c c c c}
\hline
Text prompt       & mIoU  & mACC  & mIoU-novel & pACC-novel
\\\hline
Manual            & 36.99 & 50.97 & 27.29       & 29.48      \\
Learnt            & 35.88 & 50.68 & 29.61       & 39.83      \\
\rowcolor{black!10} Learnt + \textit{Sub}PT & \textbf{37.89} & \textbf{51.36} & \textbf{32.84} & \textbf{42.13} \\
\hline
\end{tabular}
}
\vspace{.2mm}
\label{tab: zsseg}
\end{table}
\begin{table}[t]
\renewcommand\arraystretch{1.3}
\begin{center}
\caption{Ablation study on $t_{\text{early}}$ in \textit{Sub}PT. \textbf{Best} results are highlighted.}
\vspace{-1.mm}
\label{tab: ablation-SubPT-tearly}
\scalebox{0.92}{
\begin{tabular}{c | c | c | c c c c c c}
\hline
Dataset  & \!\!\!shots\!\!\! & \!\!\!epoch\!\!\! & \multicolumn{6}{c}{$t_{\text{early}}$} \\ \hline
\multirow{6}{*}{\!\!\!Caltech101\!\!\!} & \multirow{2}{*}{1} & \multirow{2}{*}{50} 
                         & 0     & 10    & 20    & 30    & 40    & 50    \\
         &       &       & \cellcolor{black!10} 87.37 & \cellcolor{black!10}88.72 & \cellcolor{black!10}88.76 & \cellcolor{black!10}\textbf{88.80} & \cellcolor{black!10}88.48 & \cellcolor{black!10}88.19 \\\cline{2-9}
         & \multirow{2}{*}{2} & \multirow{2}{*}{100} 
                         & 0     & 20    & 40    & 60    & 80    & 100    \\
         &       &       & \cellcolor{black!10}87.99 & \cellcolor{black!10}\textbf{89.52} & \cellcolor{black!10}88.46 & \cellcolor{black!10}88.15 & \cellcolor{black!10}88.03 & \cellcolor{black!10}87.79 \\ \cline{2-9}
         & \multirow{2}{*}{4} & \multirow{2}{*}{100} 
                         & 0     & 30    & 40    & 60    & 80    & 100    \\
         &       &       & \cellcolor{black!10}89.48 & \cellcolor{black!10}\textbf{90.39} & \cellcolor{black!10}90.10 & \cellcolor{black!10}89.17 & \cellcolor{black!10}89.21 & \cellcolor{black!10}89.49 \\ \hline
\multirow{6}{*}{\!\!Food 101\!\!} & \multirow{2}{*}{1} & \multirow{2}{*}{50} 
                         & 0     & 10    & 20    & 30    & 40    & 50    \\
         &       &       & \cellcolor{black!10} 74.52 & \cellcolor{black!10}\textbf{77.18} & \cellcolor{black!10}75.83 & \cellcolor{black!10}75.52 & \cellcolor{black!10}75.02 & \cellcolor{black!10}74.45 \\\cline{2-9}
         & \multirow{2}{*}{2} & \multirow{2}{*}{100} 
                         & 0     & 20    & 40    & 60    & 80    & 100    \\
         &       &       & \cellcolor{black!10}72.48 & \cellcolor{black!10}\textbf{76.34} & \cellcolor{black!10}73.69 & \cellcolor{black!10}72.93 & \cellcolor{black!10}72.58 & \cellcolor{black!10}72.28 \\ \cline{2-9}
         & \multirow{2}{*}{4} & \multirow{2}{*}{100} 
                         & 0     & 30    & 40    & 60    & 80    & 100    \\
         &       &       & \cellcolor{black!10}73.54 & \cellcolor{black!10}\textbf{77.69} & \cellcolor{black!10}77.41 & \cellcolor{black!10}76.32 & \cellcolor{black!10}75.97 & \cellcolor{black!10}75.71 \\ \hline
\end{tabular}
}
\end{center}
\end{table}
\begin{table}[t]
\renewcommand\arraystretch{1.3}
\begin{center}
\caption{Ablation study on the subspace rank $r$ in \textit{Sub}PT. \textbf{Best} results are highlighted.}
\vspace{-1.mm}
\label{tab: ablation-SubPT-rank}
\scalebox{.93}{
\begin{tabular}{c | c c c c c c}
\hline
\multirow{2}{*}{CoOp + \textit{Sub}PT}& \multicolumn{6}{c}{subspace rank $r$} \\
           & 5     & 10    & 15    & 20    & 25    & 30 \\\hline
Caltech101 & 90.22 & 90.39 & 90.26 & 90.39 & \textbf{90.55} & 90.51 \\
Food 101   & 77.62 & 77.69 & 77.93 & \textbf{78.04} & 78.02 & 78.00 \\
\hline
\end{tabular}
}
\end{center}
\end{table}
\begin{table}[t]
\renewcommand\arraystretch{1.3}
\begin{center}
\caption{Ablation study on the number of novel category names in NFL.}
\vspace{-1.mm}
\label{tab: ablation-NFL-n}
\scalebox{.93}{
\begin{tabular}{c | c c c c c}
\hline
\multirow{2}{*}{CoOp + NFL}& \multicolumn{5}{c}{number of novel category names in NFL} \\
              & 1     & 10    & 100   & 200   & 500   \\ \hline
Caltech101    & 88.32 & \textbf{88.82} & 88.80 & 88.56 & 88.69 \\
Food101       & 74.91 & 75.85 & 76.02 & 75.99 & \textbf{76.09} \\ 
\hline
\end{tabular}
}
\end{center}
\end{table}
\begin{table}[t]
\renewcommand\arraystretch{1.3}
\begin{center}
\caption{Comparison on average time cost in a single iteration under 1-shot setting.}
\vspace{-1.mm}
\label{tab: training-time-comparison}
\scalebox{.95}{
\begin{tabular}{c | c c | c}
\hline
Method  & \textit{Sub}PT & NFL        & Training time (s) \\ \hline
CoOp    &                &            & 0.19          \\
\rowcolor{black!10}CoOp    & \checkmark     &            & 0.20          \\
\rowcolor{black!10}CoOp    &                & \checkmark & 0.20          \\
\rowcolor{black!10}CoOp    & \checkmark     & \checkmark & 0.22          \\\hline
CoCoOp  &                &            & 1.40          \\\hline
\end{tabular}
}
\end{center}
\end{table}
\begin{table}[h]
\renewcommand\arraystretch{1.25}
\begin{center}
\caption{Ablation study on token number in terms of base-to-novel generalization.}
\vspace{-1.mm}
\label{tab: ablation-token-number}
\scalebox{0.94}{
\begin{tabular}{c|c|c|c c c}
\hline
Dataset                     & Method                          & \!\# Token\! & \multicolumn{1}{l}{Base}  & \multicolumn{1}{l}{Novel} & HM    \\ \hline
\multirow{9}{*}{\!\!\!Caltech101\!\!} & \multirow{3}{*}{CoOp}           & 4            & \multicolumn{1}{l}{93.78} & \multicolumn{1}{l}{86.17} & 89.81 \\
                            &                                 & 16           & 94.00                     & 87.37                     & 90.56 \\
                            &                                 & 60           & 93.22                     & 87.08                     & 90.05 \\ \cline{2-6} 
                            &\multirow{3}{*}{CoOp + \textit{Sub}PT}     & 4            & \multicolumn{1}{l}{94.34} & \multicolumn{1}{l}{89.34} & 91.77 \\
                            &                                 & 16           & 94.35                     & 89.27                     & 91.74 \\
                            &                                 & 60           & 94.25                     & 89.85                     & 92.00 \\ \cline{2-6} 
                            &\!\!\!\multirow{3}{*}{CoOp + \textit{Sub}PT + NFL\!\!\!} & 4            & \multicolumn{1}{l}{\textbf{94.79}} & \multicolumn{1}{l}{91.48} & 93.11 \\
                            &                                 & 16           & 94.77                     & \textbf{92.10}                     & \textbf{93.42} \\
                            &                                 & 60           & 94.77                     & 91.34                     & 93.02 \\ \hline
\multirow{9}{*}{\!\!Food 101\!}   & \multirow{3}{*}{CoOp}           & 4            & \multicolumn{1}{l}{76.82} & \multicolumn{1}{l}{75.42} & 76.11 \\
                            &                                 & 16           & 78.59                     & 77.86                     & 78.22 \\
                            &                                 & 60           & 74.77                     & 75.05                     & 74.91 \\ \cline{2-6} 
                            &\multirow{3}{*}{CoOp + \textit{Sub}PT}     & 4            & \multicolumn{1}{l}{79.69} & \multicolumn{1}{l}{79.44} & 79.56 \\
                            &                                 & 16           & 80.76                     & 80.85                     & 80.80 \\
                            &                                 & 60           & 79.20                     & 80.43                     & 79.81 \\ \cline{2-6} 
                            &\!\!\!\multirow{3}{*}{CoOp + \textit{Sub}PT + NFL\!\!\!} & 4            & \multicolumn{1}{l}{79.47} & \multicolumn{1}{l}{83.47} & 81.42 \\
                            &                                 & 16           & \textbf{82.20}                     & \textbf{84.14}                     & \textbf{83.16} \\
                            &                                 & 60           & 80.33                     & 83.16                     & 81.72 \\ \hline
\end{tabular}
}
\vspace{-.08cm}
\end{center}
\end{table}
\begin{table}[h]
\renewcommand\arraystretch{1.3}
\begin{center}
\caption{Ablation study on manual template in NFL. Colored rows refer to the default setting in main experiments.}
\vspace{-1.mm}
\label{tab: ablation-NFL-template}
\scalebox{0.95}{
\begin{tabular}{c | l | c c c}
\hline
Dataset                     & Method              & Base  & Novel & HM    \\ \hline
\multirow{4}{*}{Caltech101} & CoOp                             & 94.00 & 87.37 & 90.56 \\
                            & CoOp + \textit{Sub}PT + NFL (T1) & 94.73 & 91.08 & 92.92 \\
                            & CoOp + \textit{Sub}PT + NFL (T2) & \textbf{94.79} & 91.38 & 93.05 \\
                            & \cellcolor{black!10}CoOp + \textit{Sub}PT + NFL (T3) & \cellcolor{black!10}94.77 & \cellcolor{black!10}\textbf{92.10} & \cellcolor{black!10}\textbf{93.42} \\ \hline
\multirow{4}{*}{Food 101}   & CoOp                             & 78.59 & 77.86 & 78.22 \\
                            & CoOp + \textit{Sub}PT + NFL (T1) & 80.84 & 83.94 & 82.36 \\
                            & CoOp + \textit{Sub}PT + NFL (T2) & 81.87 & 83.99 & 82.92 \\
                            & \cellcolor{black!10}CoOp + \textit{Sub}PT + NFL (T3) & \cellcolor{black!10}\textbf{82.20} & \cellcolor{black!10}\textbf{84.14} & \cellcolor{black!10}\textbf{83.16} \\ \hline
\end{tabular}
}
\end{center}
\end{table}
\begin{table}[h]
\renewcommand\arraystretch{1.3}
\begin{center}
\caption{Ablation study on specified categories in NFL. Colored rows refer to the default setting in main experiments.}
\vspace{-1.mm}
\label{tab: ablation-ProGrad}
\scalebox{0.95}{
\begin{tabular}{c | l | c c c}
\hline
Dataset                     & Method                       & Base  & Novel & HM    \\ \hline
\multirow{6}{*}{Caltech101} & CoOp                         & 94.00 & 87.37 & 90.56 \\
                            & CoOp + \textit{Sub}PT        & 94.35 & 89.27 & 91.74 \\
                            & CoOp + \textit{Sub}PT + NFL (base)            & \textbf{95.09} & 89.12 & 92.01 \\
                            & \cellcolor{black!10}CoOp + \textit{Sub}PT + NFL (novel)           & \cellcolor{black!10}94.77 & \cellcolor{black!10}\textbf{92.10} & \cellcolor{black!10}\textbf{93.42} \\
                            & CoOp + \textit{Sub}PT + NFL (whole)           & 94.88 & 90.36 & 92.56 \\ 
                            & ProGrad \cite{zhu2022prompt}                  & 94.45 & 87.99 & 91.11 \\
                            \hline
\multirow{6}{*}{Food 101}   & CoOp                         & 78.59 & 77.86 & 78.22 \\
                            & CoOp + \textit{Sub}PT        & 80.76 & 80.85 & 80.80 \\
                            & CoOp + \textit{Sub}PT + NFL (base)            & \textbf{82.35} & 82.15 & 82.25 \\
                            & \cellcolor{black!10}CoOp + \textit{Sub}PT + NFL (novel)           & \cellcolor{black!10}82.20 & \cellcolor{black!10}\textbf{84.14} & \cellcolor{black!10}\textbf{83.16} \\
                            & CoOp + \textit{Sub}PT + NFL (whole)           & 82.22 & 83.56 & 82.88 \\
                            & ProGrad \cite{zhu2022prompt}                  & 81.50 & 83.23 & 82.36 \\
                            \hline
\end{tabular}
}
\end{center}
\end{table}
\begin{figure}[t]
    \centering
    \scalebox{0.34}{
    \includegraphics{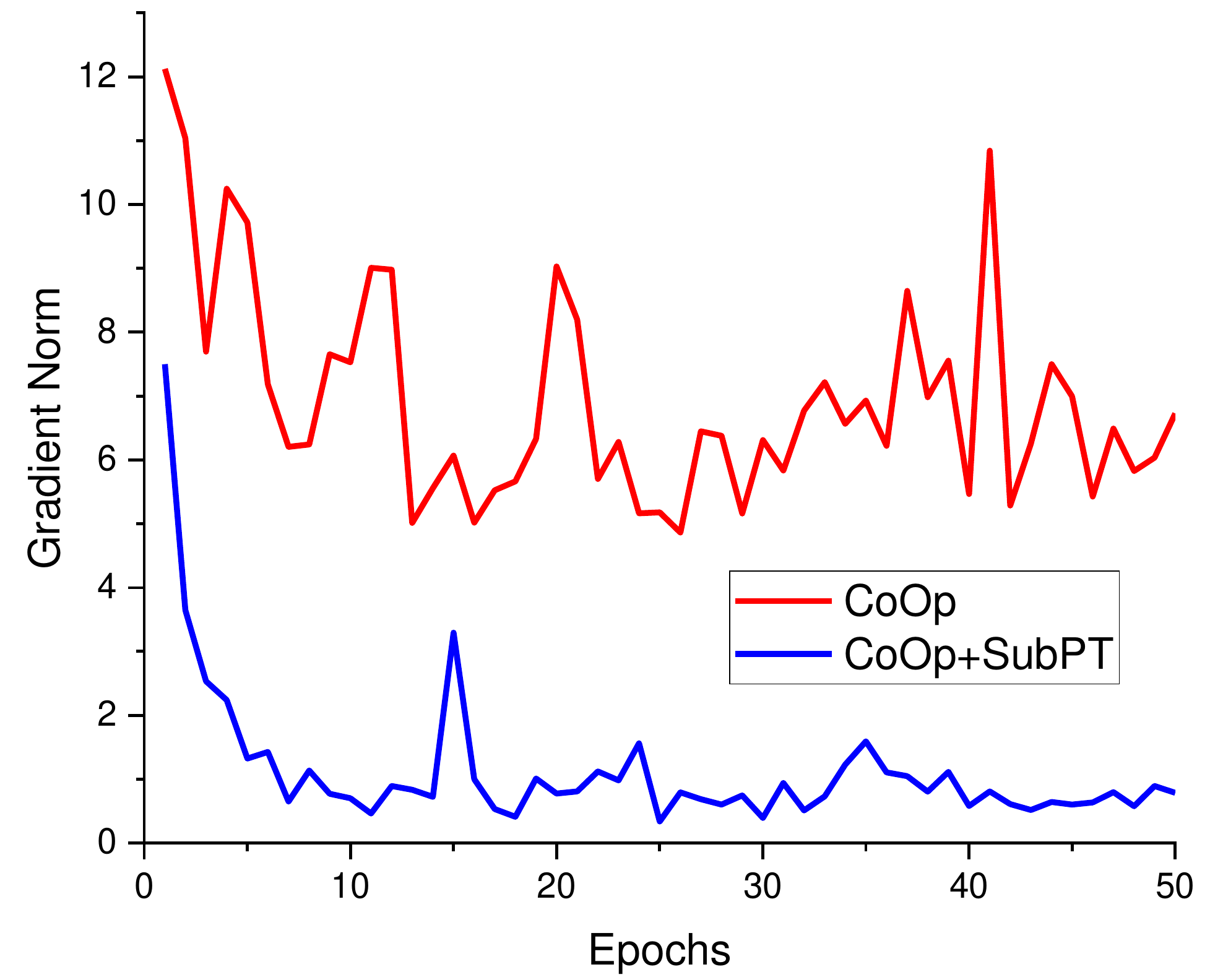}
    }
    \vspace{-1mm}
    \caption{Comparison of the gradient norm between the CoOp training and the CoOp+\textit{Sub}PT training.}
    \label{fig: gradient-norm}
\end{figure}
\begin{figure}[t]
    \centering
    \scalebox{0.45}{
    \includegraphics{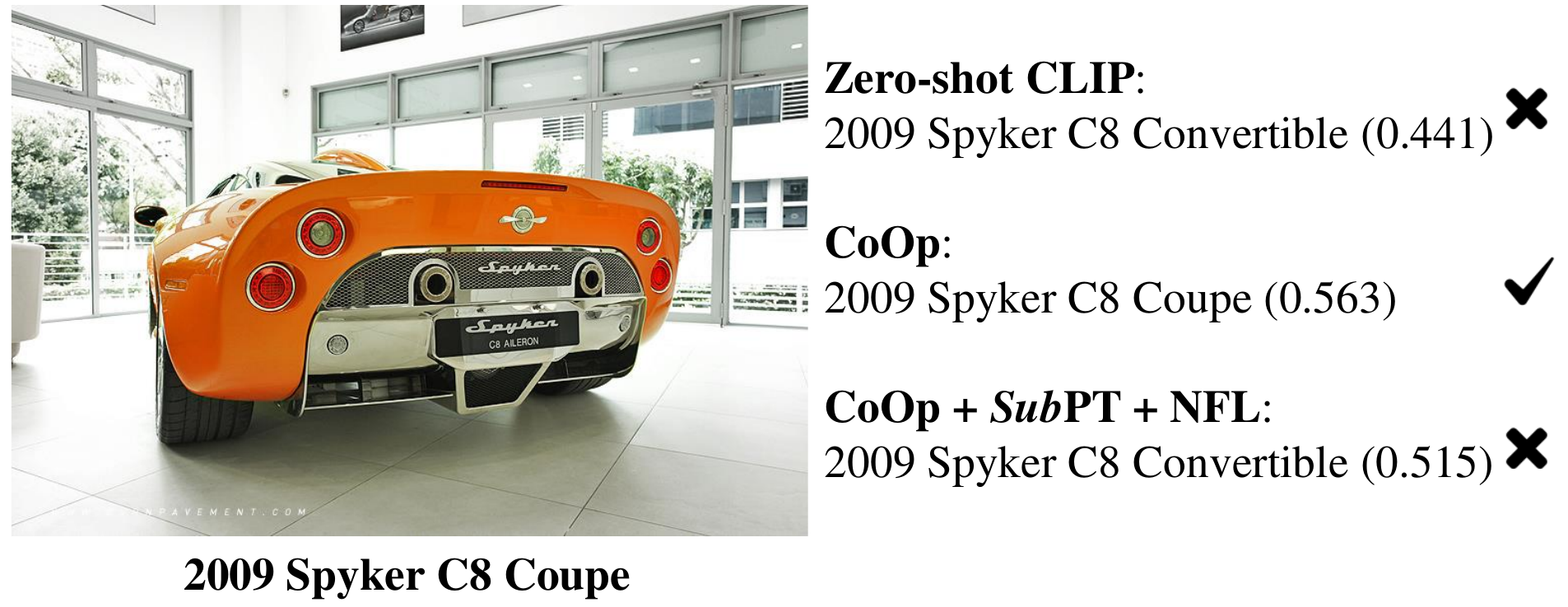}
    }
    \caption{A failure classification case. Figures in parentheses are the predicted confidence scores. Zero-shot CLIP and our method cannot classify correctly, while CoOp can instead.}
    \label{fig: failure-case}
\end{figure}

\vspace{1mm}
\noindent\textbf{Open-Vocabulary Object Detection Results.}
As reported in Table \ref{tab: owod}, RPL in the 1-shot setting can decrease the AP$_{novel}$ score from 7.4\% to 6.4\% compared with the zero-shot baseline, indicating the ineffectiveness of RPL under low-shots condition, whereas our proposed \textit{Sub}PT can boost RPL's detection performance up to 9.7\%, surpassing the zero-shot baseline.

\subsection{Zero-Shot Semantic Segmentation}
Finally, we consider the challenging downstream vision task, that is \textit{zero-shot semantic segmentation}.

\vspace{1mm}
\noindent\textbf{Datasets and Evaluation Metrics.} 
We follow the recent work \textit{ZSSEG} \cite{zsseg} and rely on the COCO stuff \cite{coco-stuff} dataset containing 171 annotated categories, further divided into 156 base categories and 15 novel categories.
We consider four evaluation metrics on 5k validation images, \ie, mean intersection over union across all categories (mIoU) and only novel categories (mIoU-novel), mean pixel accuracy averaged across all categories (mACC), and pixelwise classification accuracy on novel categories (pACC-novel).

\vspace{1mm}
\noindent\textbf{Baselines and Implementation Details.}\;\;
To conduct zero-shot semantic segmentation, we follow the pipeline proposed in \textit{ZSSEG}\footnote{\url{https://github.com/MendelXu/zsseg.baseline}} \cite{zsseg} containing two procedures: mask proposal generation and region classification via prompt-tuned CLIP.
For the latter procedure, we implement our proposed \textit{Sub}PT into conventional prompt tuning, in which we conduct 2-shots training and follow the same training settings in \textit{ZSSEG}, with batch size being 32 and base learning rate being 0.02 (adjusted by cosine annealing schedule).
The ViT-B/16 visual encoder is utilized in CLIP, and the MaskFormer \cite{maskformer} model with ResNet-50 is selected as backbone for semantic segmentation.

\vspace{1mm}
\noindent\textbf{Zero-Shot Semantic Segmentation Results.}
Table \ref{tab: zsseg} illustrates that learnable prompts can improve the classification accuracy on image pixels within novel categories (mIoU-novel and pACC-novel) compared with manual prompts, whereas our \textit{Sub}PT can further enhance the performances with large margins, 32.84\% \textit{vs} 29.61\% on mIoU-novel, for example.
These results verify the effectiveness of \textit{Sub}PT across various downstream tasks.

\subsection{Ablative Study}
In this section, we ablate all the hyper-parameters in the proposed method. Moreover, we compare the training time among baseline methods and our proposed method.

\vspace{1mm}
\noindent\textbf{Effect of $t_{\text{early}}$ on \textit{Sub}PT.}\;\;\;
\textit{Sub}PT only has two hyper-parameters, namely $t_{\text{early}}$ and $r$, in which $t_{\text{early}}$ denotes the last epoch of the early training stage.
As presented in Sec. \ref{sec: Cause of overfitting phenomenon}, we propose to eliminate the overfitting problem by multiplying the gradient by eigenvectors representing the early-stage gradient flow during back-propagation. 
As $t_{\text{early}}$ defines the range of early training stage, it can decide the quality of computed $\boldsymbol{U}^{\text{early}}$ and further influence the effect of \textit{Sub}PT.
Here, $t_{\text{early}}=0$ indicates the original CoOp for comparison.
As depicted in Table \ref{tab: ablation-SubPT-tearly}, a larger $t_{\text{early}}$ leads to less satisfying few-shot classification performance.
When $t_{\text{early}}$ equals the total epoch, the performance approximates that of the original CoOp, indicating that the computed eigenvectors $\boldsymbol{U}^{\text{early}}$ represent the gradient flow of the entire training process and lose the ability to mitigate overfitting.

\vspace{1mm}
\noindent\textbf{Effect of subspace rank $r$ on \textit{Sub}PT.}
We investigate how the subspace rank $r$, which is the number of eigenvectors in $\boldsymbol{U}^{\text{early}}$, affects the performance of \textit{Sub}PT.
As shown in Table \ref{tab: ablation-SubPT-rank}, the performance of CoOp+\textit{Sub}PT on few-shot classification is robust across different values of subspace rank $r$.
This scenario can be explained by the spectrum of gradient flow being dominated by only a few components, as presented in Sec. \ref{sec: Analyzing gradient flow in prompt tuning}.

\vspace{1mm}
\noindent\textbf{Effect of the number of novel category names on NFL.}
Table \ref{tab: ablation-NFL-n} shows the effect of the number of novel category names in NFL. Increasing the number from 100 to 500 will not bring obvious performance improvements. Thus, we fix the number $n$ as 100 by default if it is unspecified.

\vspace{1mm}
\noindent\textbf{Computational efficiency.}\;\;
We compare the average time cost in a single iteration under 1-shot setting over the Caltech101 dataset, trained on an Nvidia RTX A6000 GPU. We observe from Table \ref{tab: training-time-comparison} that both \textit{Sub}PT and NFL will not burden the efficiency of CoOp and are much more effective than CoCoOp (0.22s \textit{vs} 1.40s).

\vspace{1mm}
\noindent\textbf{Quantitative analysis of the gradient norm.}
As \textit{Sub}PT eliminates the spurious components in gradients during back-propagation, it is reasonable for the gradient norm in the later training stage to be smaller than that in the early stage.
We record the gradient norms during the entire training process and observe the corresponding results, as shown in Fig. \ref{fig: gradient-norm}.

\vspace{1mm}
\noindent\textbf{Effect of token number.}\;\;\;\; 
We further investigate how the number of learnable word embeddings affects the performances on both base and novel categories. We repeat the base-to-novel generalization experiments with different token number $M\in\{4, 16, 60\}$.
As shown in Table {\ref{tab: ablation-token-number}}, the performances on both base and novel categories slightly drop within 1.5\% with $M$ not being 16 (default), and this trend always holds among different methods.

\vspace{1mm}
\noindent\textbf{Effect of different manual templates on NFL.}\;\; 
We repeat the base-to-novel generalization experiments with three different manual templates T1/T2/T3. 
T1 is the classical one ``a photo of a [CLASS]". 
T2 is the ensemble of 7 selected handcrafted provided by \cite{CoOp}, including ``itap of a [CLASS]", ``a bad photo of the [CLASS]", ``a origami [CLASS]", ``a photo of the large [CLASS]", ``a [CLASS] in a video game", ``art of the [CLASS]", and ``a photo of the small [CLASS]".
T3 is the ensemble of 80 handcrafted templates provided by \cite{CLIP}, which is the default setting in this paper.
As shown in Table {\ref{tab: ablation-NFL-template}}, T3 can slightly boost the performances on both base and novel categories with approximately 1\% margins compared with T1.

\subsection{Further Analysis}
\noindent\textbf{Effect of NFL on base/novel/whole categories}.\;\;\; We provide more discussions on the design of NFL. We attempt to regularize the feature similarity on base/novel/whole categories respectively, and the base-to-novel performances are shown in Table \ref{tab: ablation-ProGrad}.

Clearly, NFL (novel), our default setting in the main experiments, can always contribute to the best novel accuracy and the best harmonic mean accuracy among all methods. The reason lies in the ability of NFL (novel) to directly optimize novel categories. Although the base/novel trade-off in NFL (novel) slightly affects the base accuracy, it can still outperform the CoOp baseline and CoOp+\textit{Sub}PT by large margins (\eg, 82.20\% \textit{vs} 78.59\% and 80.76\% on the Food101 dataset).

Furthermore, NFL (base) can also enhance the base and novel performances. This finding is reasonable, as regularization on base categories can prevent CoOp training from forgetting the general knowledge learned from zero-shot CLIP and further mitigate overfitting. Such a point is also considered by the concurrent work ProGrad \cite{zhu2022prompt}, which conducts knowledge distillation on the predicted softmax scores for base categories. For completeness, we also make an experimental comparison with ProGrad in Table \ref{tab: ablation-ProGrad}. 

The performances of NFL (whole) are between those of NFL (base) and NFL (novel).
    
\noindent\textbf{Failure cases.}\;\;
As our \textit{Sub}PT strategy is designed for mitigating overfitting, its performance improvements will be not obvious enough for some rare datasets in which the vanilla CoOp does not encounter the overfitting problem. 
For example, under the 1-shot setting, CoOp+\textit{Sub}PT can boost the base class accuracy over CoOp with only a 0.33\% absolute margin over the Flowers 102 dataset (66.14\% \textit{vs} 65.81\%).

Regarding NFL, Fig. {\ref{fig: failure-case}} shows a failure case where both zero-shot CLIP and our CoOp+\textit{Sub}PT+NFL fail to classify the test image correctly, but CoOp can achieve the opposite.
The reason behind this is that the NFL term introduces an incorrect supervision signal from zero-shot CLIP and misleads the CoOp training. 
Such a trend can also be observed on a few datasets such as DTD and UCF 101, under the base-to-novel generalization setting (see Table \ref{tab: base2novel}).
Despite this finding, NFL can still largely boost the novel class accuracies and finally lead to better harmonic mean accuracies over the vanilla CoOp, which enhances the generalization capability of learned prompts.

\section{Conclusion}
In this research, we first revisit two aspects of overfitting problem appeared in the well-known prompt tuning approach CoOp. 
Then, we explore the cause of overfitting by measuring the gradient flow, and observe through experiments that CoOp learns orthogonal features at the early and later training stage, which leads to the non-overfitting and overfitting phenomenon, respectively.
{On this basis, we propose Subspace Prompt Tuning (\textit{Sub}PT). During the whole training process, \textit{Sub}PT projects the gradients in back-propagation onto the low-rank subspace spanned by the eigenvectors representing the early-stage gradient flow, and successfully mitigate the overfitting problem.
In addition, we equip CoOp with the specially designed Novel Feature Learner (NFL) to improve the generalization ability of the learned prompt towards the novel classes out of the training set, needless of image training data.
Extensive experiments on image classification, open-vocabulary object detection, and zero-shot semantic segmentation verify the effectiveness of the proposed method.
Our future work will include designing a more robust and general subspace for multimodal prompt tuning.

\bibliographystyle{IEEEtran}
\bibliography{reference_bibtex}

\end{document}